\documentclass[lettersize,journal]{IEEEtran}
\usepackage{amsmath,amsfonts}
\usepackage{gensymb}
\usepackage{algorithmic}
\usepackage{algorithm}
\usepackage{array}
\usepackage[caption=false,font=normalsize,labelfont=sf,textfont=sf]{subfig}
\usepackage{textcomp}
\usepackage{stfloats}
\usepackage{url}
\usepackage{verbatim}
\usepackage{graphicx}
\usepackage{cite}
\usepackage{multirow}
\usepackage{enumerate}
\usepackage{booktabs}
\usepackage{hyperref}
\usepackage{marvosym}
\hypersetup{
    colorlinks=true,            
    linkcolor=black,             
    filecolor=magenta,          
    urlcolor=black,              
    pdftitle={Overleaf Example},
    pdfpagemode=FullScreen,
    }
\usepackage{color}
\definecolor{ourred}{RGB}{200,36,35}
\definecolor{ourblue}{RGB}{12,132,198}

\begin{document}

\title{A Surrogate-Assisted Extended Generative Adversarial Network for Parameter Optimization in Free-Form Metasurface Design}

\author{Manna Dai, Yang Jiang,~\IEEEmembership{Member,~IEEE,} Feng Yang, Joyjit Chattoraj, Yingzhi Xia, Xinxing Xu,~\IEEEmembership{Member,~IEEE,} Weijiang Zhao, My Ha Dao, and Yong Liu
\thanks{Manuscript received October 12, 2023. This work was financially supported by the A*STAR AME Programmatic Fund (Grant No. A20H5b0142). \textit{(Corresponding author: Yong Liu)}}
\thanks{Manna Dai, Feng Yang, Joyjit Chattoraj, Yingzhi Xia, Xinxing Xu, and Yong Liu are with the Computing and Intelligence Department, Institute of High Performance Computing (IHPC), Agency for Science, Technology and Research (A*STAR), 1 Fusionopolis Way, \#16-16 Connexis, 138632, Republic of Singapore (e-mail: \href{mailto:manna\_dai@ihpc.a-star.edu.sg}{manna\_dai@ihpc.a-star.edu.sg}; \href{mailto:yangf@ihpc.a-star.edu.sg}{yangf@ihpc.a-star.edu.sg}; \href{mailto:joyjit\_chattoraj@ihpc.a-star.edu.sg}{joyjit\_chattoraj@ihpc.a-star.edu.sg}; \href{mailto:xia\_yingzhi@ihpc.a-star.edu.sg}{xia\_yingzhi@ihpc.a-star.edu.sg}; \href{mailto:xuxinx@ihpc.a-star.edu.sg}{xuxinx@ihpc.a-star.edu.sg}; \href{mailto:liuyong@ihpc.a-star.edu.sg}{liuyong@ihpc.a-star.edu.sg}).} 
\thanks{Yang Jiang and Weijiang Zhao are with the Electronics and Photonics Department, Institute of High Performance Computing (IHPC), Agency for Science, Technology and Research (A*STAR), 1 Fusionopolis Way, \#16-16 Connexis, 138632, Republic of Singapore (e-mail:\href{mailto:jiang\_yang@ihpc.a-star.edu.sg}{jiang\_yang@ihpc.a-star.edu.sg}; \href{mailto:zhaow@ihpc.a-star.edu.sg}{zhaow@ihpc.a-star.edu.sg}).} 
\thanks{My Ha Dao is with the Fluid Dynamics Department, Institute of High Performance Computing (IHPC), Agency for Science, Technology and Research (A*STAR), 1 Fusionopolis Way, \#16-16 Connexis, 138632, Republic of Singapore (e-mail:\href{mailto:daomh@ihpc.a-star.edu.sg}{daomh@ihpc.a-star.edu.sg}).} }



\maketitle

\begin{abstract}
Metasurfaces have widespread applications in fifth-generation (5G) microwave communication. Among the metasurface family, free-form metasurfaces excel in achieving intricate spectral responses compared to regular-shape counterparts. However, conventional numerical methods for free-form metasurfaces are time-consuming and demand specialized expertise. Alternatively, recent studies demonstrate that deep learning has great potential to accelerate and refine metasurface designs. Here, we present XGAN, an extended generative adversarial network (GAN) with a surrogate for high-quality free-form metasurface designs. The proposed surrogate provides a physical constraint to XGAN so that XGAN can accurately generate metasurfaces monolithically from input spectral responses. In comparative experiments involving 20000 free-form metasurface designs, XGAN achieves 0.9734 average accuracy and is 500 times faster than the conventional methodology. This method facilitates the metasurface library building for specific spectral responses and can be extended to various inverse design problems, including optical metamaterials, nanophotonic devices, and drug discovery.
\end{abstract}


\section{Introduction}
\IEEEPARstart{T}{his} Metasurfaces, as the two-dimensional (2D) counterparts of metamaterials, enable the manipulation of electromagnetic (EM) behaviors, such as phases, amplitudes, and polarization of reflected/transmitted waves in space \cite{yang2021demonstration,cheng2023large,shaltout2019spatiotemporal}. With the advent of fifth-generation (5G) wireless communication, metasurfaces have garnered considerable scientific and engineering interest due to the advantages of planar profile, easy integration, and low loss \cite{chen2020microwave}. The concept of ``digital metamaterials" allowed for the coding of metasurfaces to achieve different EM responses \cite{cui2014coding,wan2016field}. The conventional metasurface design involves three steps: determining metasurface patterns by a metamaterial specialist, achieving the EM responses of patterns with the aid of full-wave EM simulators, and selecting the metasurface with desired EM responses. Nevertheless, the pattern design requires the expertise of a specialist. Moreover, the metasurface selection relies on trial and error \cite{qiu2019deep}, requiring tremendous iterative EM simulations to achieve a metasurface with a satisfactory spectral response. Consequently, there is a pressing need to quickly predict the EM responses of various digital patterns to facilitate rapid metasurfaces design \cite{zhu2021phase,li2017electromagnetic}. 

Deep learning \cite{jia2023knowledge} has replaced a lot of manual work in metasurface design, such as metasurface coding \cite{liu2022programmable} and real-time response simulation \cite{zhu2023metasurfaces}. The metasurface design involves forward modeling and inverse design \cite{zhang2022heterogeneous,jia2022situ}. Forward modeling, also known as surrogate modeling, predicts EM response without EM simulation by using structural patterns as input \cite{nadell2019deep,li2022empowering}. On the other hand, inverse design quickly produces structural patterns with the input of EM response \cite{tanriover2022deep,wang2022deep}. This on-demand design method overcomes the limitations of traditional methods, which are generally time-consuming, inefficient, and experience-dependent. Each metasurface design exhibits a different degree of freedom (DOF), comprising input DOF, parameter DOF, and output DOF \cite{xiong2019controlling}. Our study focuses on the parameter DOF, as it aims to control the geometric parameters of complex metasurface structures to realize desired EM responses. Liu et al. \cite{liu2021tackling} introduced deep learning as a substitute to solve the inverse design problem in photonics and optics, particularly for structure designs with high parameter DOF that are too intricate for traditional methods to handle. Deep learning models for inverse design can be categorized into discriminative models and generative models. Discriminative models capture the relationship between structural patterns and optical responses, but cannot perfectly map an optical response to a unique set of design parameters due to multiple configurations of patterns corresponding to the same response. Thus, discriminative models are suitable for low DOF structure designs and can be implemented with fundamental network architectures such as fully convolutional networks (FCN) \cite{zhang2021deep,yun2022deep} or convolutional neural networks (CNN) \cite{mall2020cyclical,hodge2019multi}. As DOF continues growing to thousands and more, generative models become useful in reducing the design's dimensionality and seeking relationships between structural patterns and optical responses for optimization. Generative adversarial networks (GANs) \cite{creswell2018generative,wang2021generative} and variational autoencoders (VAEs) \cite{kingma2014stochastic} are two commonly used deep generative models that capture the distribution of high-dimensional data and represent it in a reduced-dimensional latent space. The generative models can be jointly leveraged with discriminative models and optimization algorithms to accelerate the design process and locate global optimal solutions. 

Some studies have demonstrated the efficacy of using deep learning for synthesizing metasurfaces with extra parameter DOF. Liu et al. \cite{liu2018generative} employed a GAN to produce patterns with arbitrary topology for desired input responses. They focused on the basic shapes of binary images, including the classes of L-shape, circle, arc, square, rectangle, ellipse, and sector. Ma et al. \cite{ma2019probabilistic} utilized VAEs to encode metamaterial patterns and optical responses into a shared latent space, allowing for the automatic clustering of similar patterns and responses. Candidate patterns were generated by sampling the latent space given requirements in the decoding process. Naseri and Hum \cite{naseri2020machine} used a VAE to encode multilayer EM metasurfaces into latent vectors, which were searched using particle swarm optimization (PSO) \cite{clerc2010particle} to find an optimum feature that closely matched the desired transmission coefficient. The obtained feature was then decoded by the VAE to obtain the metasurface design. An et al. \cite{an2021multifunctional} presented a GAN that generated metasurfaces to meet multifunctional design goals, producing free-form structures in a single design iteration instead of using iterative optimization. Dai et al. \cite{dai2022slmgan} combined a GAN with a CNN-based forward model, to design free-form patterns with binary images, which could ensure that the spectral responses of generated patterns resembled the given input ones.

Although these deep learning-assisted metasurface design methods emerge in the past few years, there exist the following limitations to be resolved. Firstly, most networks tackle some simple typical geometries with limited parameter DOF, such as square \cite{liu2018generative}, H-shape \cite{ma2019probabilistic}, and circle \cite{an2020deep}. These simple geometries impose constraints on the capacity of metasurfaces to effectively produce desired responses \cite{yang2022ultraspectral}. Therefore, the generation of free-form geometries is necessary for providing high parameter DOF. Secondly, some networks produce metasurface designs that are memorized rather than generating novel solutions by learning the underlying design principles. Inverse design, as a one-to-many problem, can result in multiple patterns corresponding to a specific response. Thirdly, based on research by Yla-Oijala et al. \cite{yla2003calculation}, achieving calculation convergence for the vertex-to-vertex structure in binary matrices requires a large condition number. However, existing networks often have insufficient condition parameters, leading to numerical calculation uncertainty.

To address the above limitations, we conceptually propose an extended generative adversarial network guided by a pretrained surrogate, called XGAN. This network leverages a GAN for automatically designing free-form metasurface patterns that possess desired spectral responses. We also design a ternary coding strategy to mitigate the impact of vertex-to-vertex structure on calculation convergence. We treat the surrogate modeling as a 2D image encoding process that produces 1D features. To perform this task, we adopt a modified ResNet \cite{He2015} with a multi-head attention model \cite{vaswani2017attention}, inspired by the success of contrastive language-image pretraining (CLIP) \cite{radford2021learning} for image-to-text mapping. The resulting model is referred to as forward-ResNet (F-ResNet). We train the F-ResNet alongside a GAN to generate free-form ternary metasurface patterns capable of manipulating gigahertz (GHz) waves, which helps mitigate non-uniqueness issues in metasurface inverse designs. With GAN's rapid sampling capability, our XGAN model is competent to generate a thousand independent patterns in a second, with an accuracy of 0.9734. In this study, we developed evaluation metrics to assess XGAN, namely model fitting metrics. Our comparative analysis shows that XGAN is exceptionally effective for metasurface inverse designs. Generating high-quality patterns that match desired responses is crucial in fields like optics, electronics, and materials science, especially in the context of 5G millimeter-wave chip integration regarding frequency-selective performance. XGAN holds tremendous potential to significantly impact these fields by delivering high-quality, easy-to-manufactured high-DOF metasurface variants. 

The contributions of this work are summarized as follows.
\begin{enumerate}[1)]
	\item Pioneer the use of image-to-text encoders to create a surrogate model capable of predicting the 1D spectral responses of 2D metasurface patterns.
	\item Innovatively leverage deep learning techniques to design ternary metasurface patterns, effectively mitigating the challenges posed by vertex-to-vertex structures and enhancing calculation convergence.
	\item Design the surrogate model F-ResNet, which outperforms other CNN-based methods, offering precise response predictions and maintaining adherence to essential physical constraints during the tuning of XGAN.
	\item Propose the XGAN inverse model, which achieves a remarkable 500-fold speed improvement over traditional optimization methods like simulated annealing (SA). XGAN produces high-quality metasurface patterns that precisely match the desired spectral responses, when compared to alternative deep learning-based inverse models. 
	\item XGAN can create new solutions based on learned design principles, overcoming the drawback of tandem networks that rely on memorized metasurface designs.
\end{enumerate}

We start the rest of this paper by first reviewing the related work in Section \ref{Related Work}. Then, the methodology of XGAN is elaborated in Section \ref{Methodology}. Comparative experiments and evaluation results are presented in Section \ref{Experiment}.

\section{Related Work}\label{Related Work}

\subsection{Metasurface Inverse Design}
The inverse problem of deducing the geometric structure of a metasurface from a desired spectral response, is not easy to solve \cite{nadell2019deep}. First, when a designer specifies a particular spectral response, it may not always be feasible to find metasurface patterns capable of achieving this response. Second, there is the one-to-many mapping problem \cite{jing2022neural,tanriover2020physics}, meaning that many highly dissimilar geometric structures can yield remarkably similar spectral responses. Therefore, for a given set of desired spectral responses, there isn't a unique solution, but rather multiple feasible metasurface designs that can achieve the same goal. This challenge arises because metasurfaces with different geometric structures can be designed by various sets of physical parameters. An ideal solution to this inverse problem should address two critical aspects: (i) ascertain the existence of any valid solution and (ii) provide a set of designs that best approximate the specified response. Naseri et al. \cite{naseri2021generative} utilized a response prediction tool to serve as a surrogate models in the analysis and optimization of non-uniform metasurfaces. The integration of machine learning-based surrogate models with inverse models offers a promising solution to the one-to-many inverse design problem \cite{li2022empowering}.

\subsection{Image-to-Text Mapping: CLIP}

An image-to-text encoder \cite{zelaszczyk2023cross} is a type of neural network or model that takes an image as input and converts it into a textual representation or embedding. This technology is often used in the field of computer vision and natural language processing to bridge the gap between visual and textual information. It allows computers to understand and process images in a way that's more compatible with natural language processing.
Applications of image-to-text encoders include: 1) Image Captioning: Generating natural language descriptions for images \cite{ramos2023smallcap}. 2) Image Retrieval: Searching for images using textual queries \cite{couairon2022embedding}. 3) Object Recognition: Identifying objects and their attributes in images \cite{yao2023detclipv2}. 4) Visual Question Answering: Answering questions about images with natural language responses \cite{gao2022transform}. 5) Scene Understanding: Describing the scene, context, and relationships in an image using text \cite{kil2023prestu}. Prominent models that perform image-to-text encoding include CLIP, where both images and text are processed in a shared model to understand their relationships. These models have wide-ranging applications in fields like image classification, image generation from textual descriptions, recommendation systems, and more.

Although deep learning methods have been adopted in the area of metasurface design, to our knowledge, no researchers have attempted to emply image-to-text transformation in metasurface inverse design. As a spectral response can be regarded as a special language or text to describe a metasurface pattern, we will demonstrate that image-to-text approach can be used to automatically realize the response prediction for a given metasurface pattern.

\section{Methodology}\label{Methodology}

\subsection{Overview}

There are two components in the metasurface design framework: one
is the surrogate model (F-ResNet) that predict spectral responses from metasurface patterns, and the other is the inverse model (XGAN) that automatically generates patterns with desired spectral responses.

\subsection{Surrogate Model: F-ResNet}

\begin{figure*}[htbp]
\centering
  \includegraphics[width=0.835\linewidth]{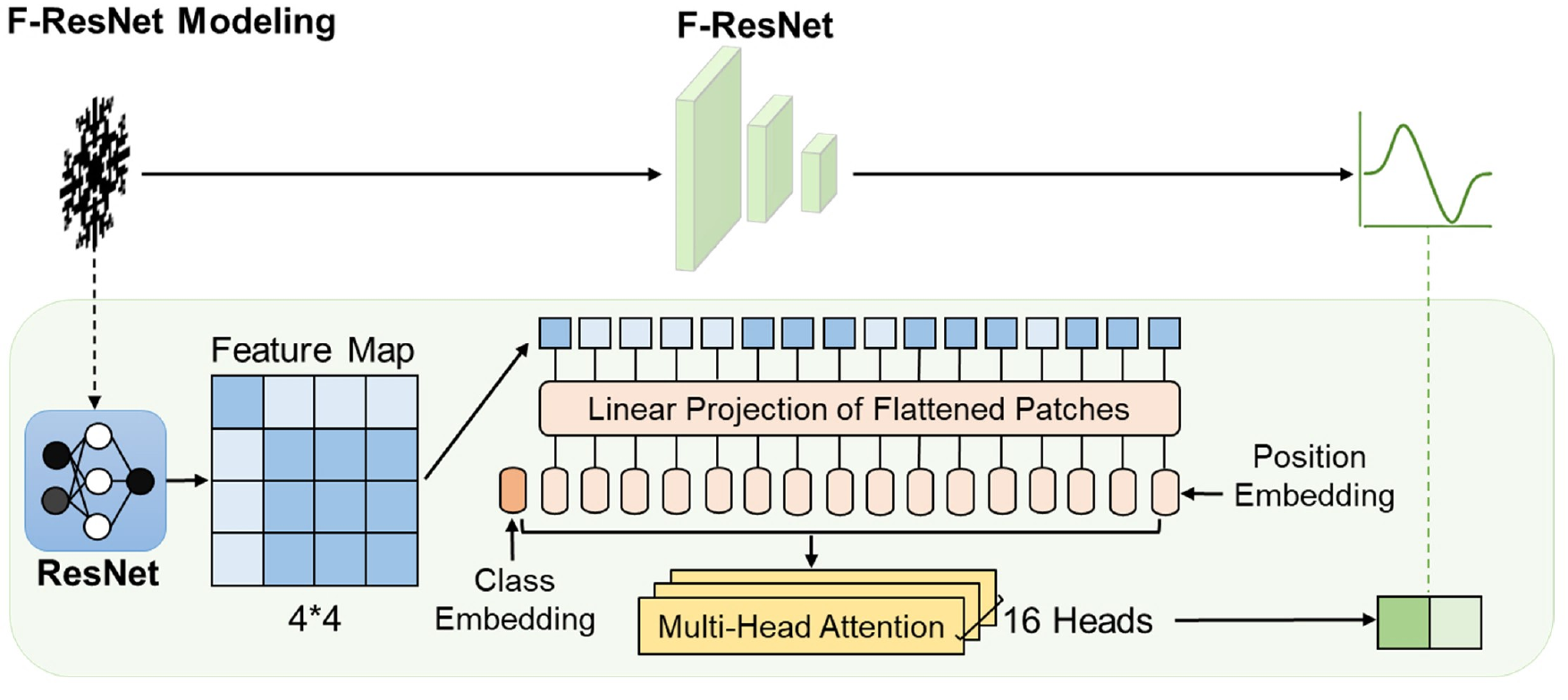}
  \caption{The network of the forward/surrogate model F-ResNet.}
  \label{fig:model_F-ResNet}
\end{figure*}

The framework of F-ResNet is depicted in Fig. \ref{fig:model_F-ResNet}. We approach the forward processing as a 1D feature extraction task applied to a 2D image. Taking inspiration from the impressive performance of CLIP in image-to-text tasks, we have leveraged its image encoder as the foundation for our surrogate model. The first step in our model is to employ ResNet50, a widely used CNN, to extract a feature map from the input image. This feature map captures high-level visual information, and then we flatten the feature map via linear projection. To generate the class embedding vector, we compute the mean of the flattened feature map. This process contributes to the improvement in model understanding of the image content and category. To provide positional information, we introduce a learnable 1D position embedding vector. This position embedding is concatenated with the class embedding as a new feature vector, enabling the model to capture spatial relationships within the image. Next, we pass the augmented feature vector through a multi-head attention model with 16 heads. This attention mechanism enhances the model's capability to encode complex sequences and capture semantic representations. Each attention head simultaneously attends to every part of the feature map, allowing for comprehensive and fine-grained analysis. Ultimately, the output of the surrogate model is considered as the extracted image feature, which serves as the response for the input image. This surrogate model works as a EM simulator, enabling us to accelerate the prediction of EM responses. Fig. \ref{fig:model_F-ResNet} provides a visual representation of the architecture of our surrogate model. During training, the loss function of our F-ResNet employs $L_{1}$ loss:

\begin{equation}\label{Eq:1}
  L_{forward}(c,c') = L_{1}(c,c')
\end{equation}
where $c$ and $c'$ represent the desired response and the generated response.

\subsection{Inverse Model: XGAN}

\begin{figure*}[htbp]
\centering
  \includegraphics[width=0.7\linewidth]{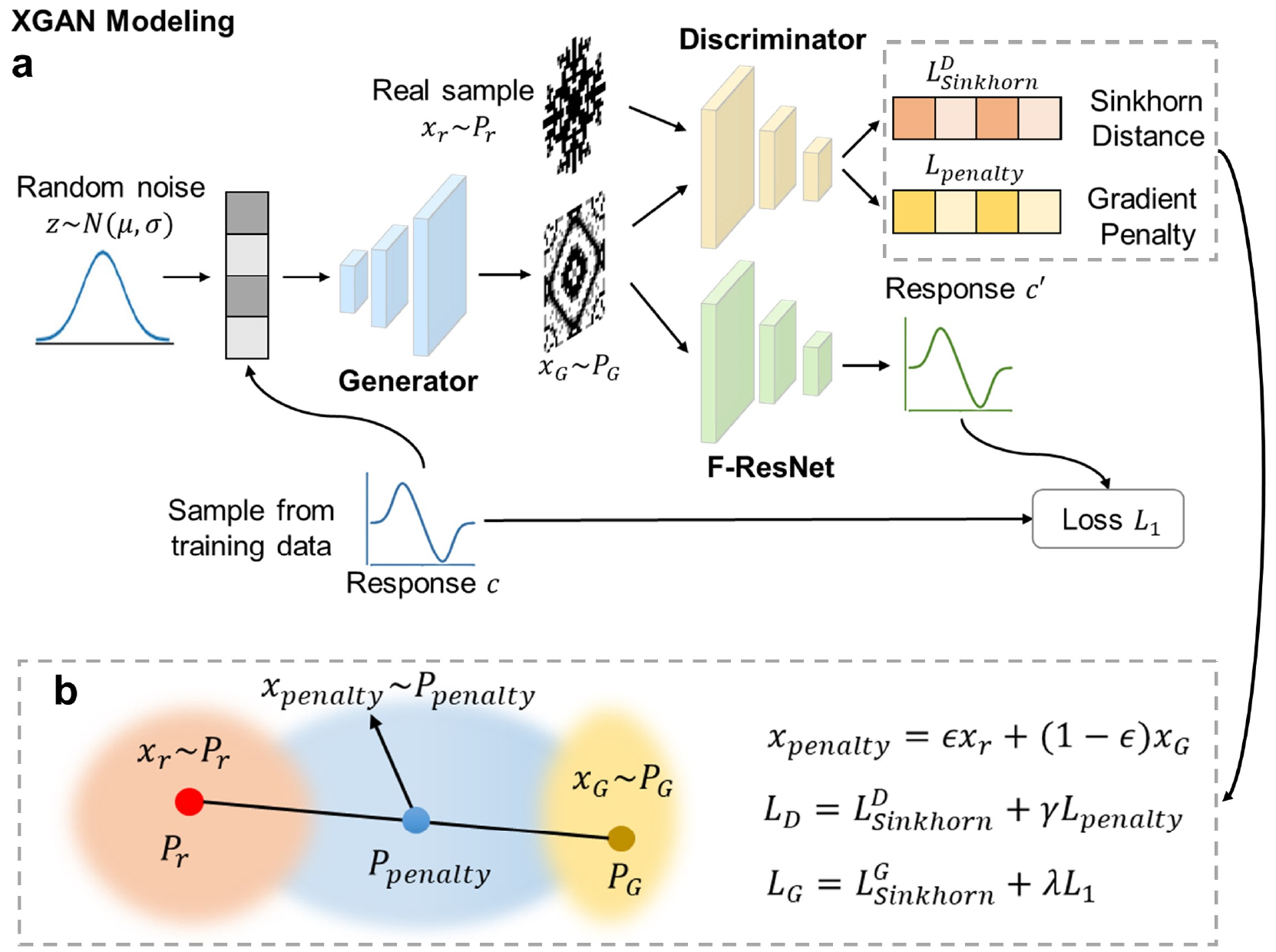}
  \caption{The modeling of XGAN. (a) The training processes of the generator and discriminator of XGAN. (b) The loss functions of XGAN consist of the discriminator loss $L_{D}$ and the generator loss $L_{G}$.}
  \label{fig:model_XGAN}
\end{figure*}

From different types of generative models, we choose GANs due to their ability to generate high-quality samples and their fast sampling capabilities. The learning process of a GAN involves an adversarial game between two neural networks, a generator, and a discriminator (Fig. \ref{fig:model_XGAN}a). During the training, the response $c$ sampled from the training data is concatenated with a random noise $z$ sampled from the Gaussian distribution $N(\mu, \sigma)$. This concatenated vector is fed into the generator and outputs a pattern $x_{G}$ which follows the distribution $P_{G}$. When the generator is trained, $x_{G}$ is fed into the discriminator and F-ResNet, and their outputs are utilized to calculate the Sinkhorn distance $L_{Sinkhorn}^{G}$ and response loss $L_{1}$, respectively. When we train the discriminator, $x_{G}$ and $x_{r}$ are fed into the discriminator to compute the Sinkhorn distance $L_{Sinkhorn}^{D}$. To deal with the issue of exploding and vanishing gradients, we gain a sample $x_{penalty}$ from the distribution $P_{penalty}$ to calculate the gradient penalty $L_{penalty}$\cite{gulrajani2017improved}. As depicted in Fig. \ref{fig:model_XGAN}b, the sample $x_{penalty}$ is calculated by the linear interpolation of $x_{r}$ and $x_{G}$.

\begin{figure*}[htbp]
\centering
  \includegraphics[width=0.7\linewidth]{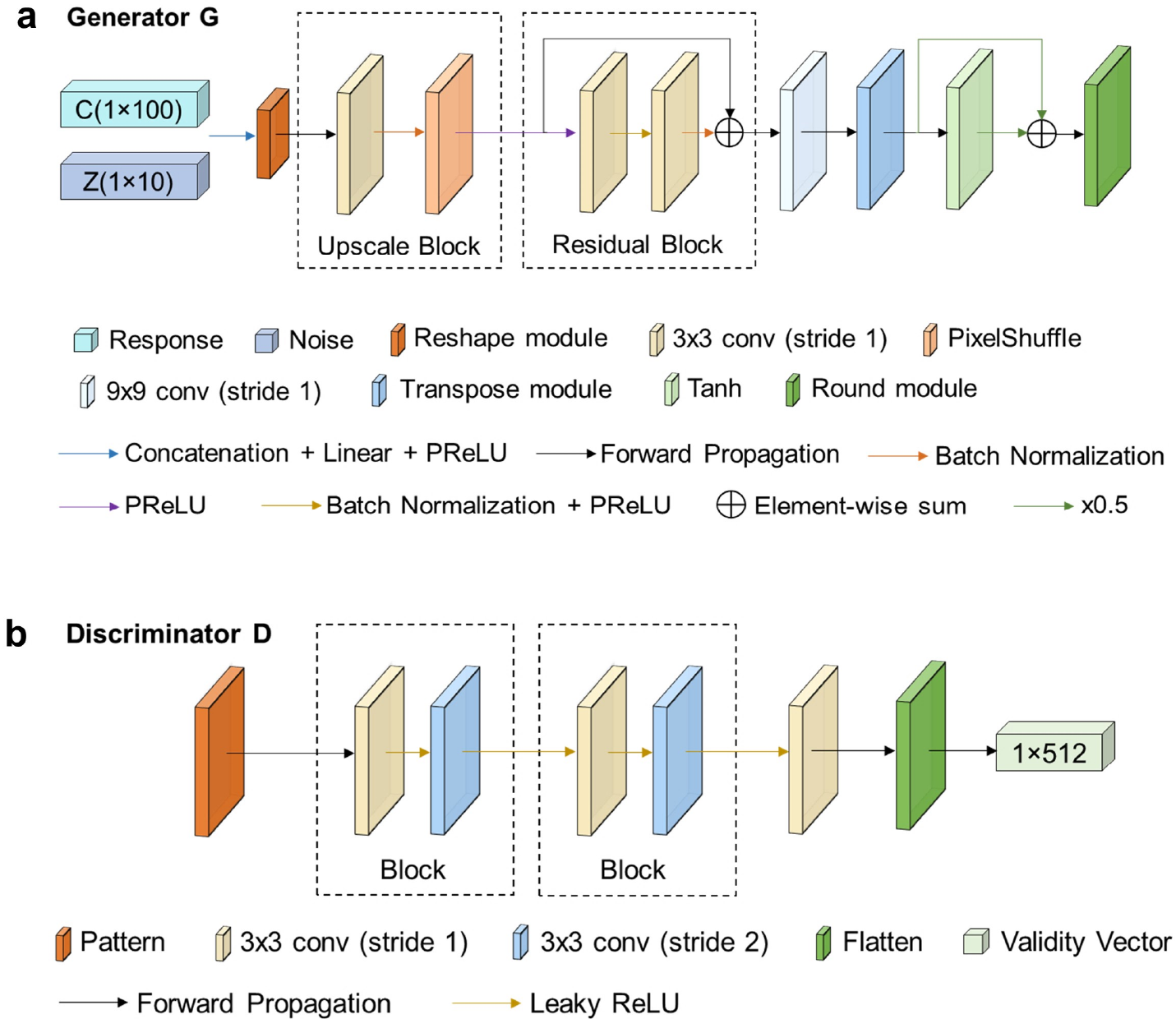}
  \caption{Proposed XGAN architecture consists of (a) a generator and (b) a discriminator.}
  \label{fig:G_D}
\end{figure*}

\subsubsection{Generator Network G}
The detailed architecture of Generator is depicted in Fig. \ref{fig:G_D}a. In the construction phase, the generator follows a specific sequence of layers and operations. This sequence involves generating patterns from a desired response vector of size $1\times 100$ and random Gaussian noise of size $1\times 10$. The construction process includes a linear layer, a parametric rectified linear unit (PReLU) layer \cite{he2015delving} for introducing nonlinear fitting, and a reshape layer to create a 2D square matrix. To enhance the resolution to a $16\times 16$ half-pattern size, an upscale block is employed, which consists of a convolutional layer, a batch-normalization layer, a pixel-shuffle layer \cite{shi2016real}, and a PReLU. The pixel-shuffle layer is responsible for upsampling the input, allowing for higher resolution and finer details in the synthesized pattern. 

To further refine the pattern synthesis, a residual block is incorporated. This block is composed of a sequence of layers: a convolutional layer, a batch-normalization layer, a PReLU layer, another convolutional layer, and another batch-normalization layer. Notably, this residual block incorporates a skip connection of ResNet50, connecting the input and output layers to alleviate the gradient vanishing problem. Following the residual block, the output passes through a convolutional layer and a Tanh activation layer, limiting the output values to the range of $[-1, 1]$. To achieve a symmetric structure, the output of the Tanh layer is added to its transposed matrix. The resulting matrix is then multiplied by 0.5 to control the output matrix values within $[-1, 1]$. Finally, a rounding function is applied to produce a ternary matrix representation of the generated image, with values $\{0, 0.5, 1\}$. By following this specific sequence of layers and operations, the generator can synthesize high-quality patterns that closely align with the desired response vector while incorporating random noise for added variation and diversity.

\subsubsection{Discriminator Network D}

In the discriminator architecture (see Fig. \ref{fig:G_D}b), the ternary pattern is provided as input. The input pattern is then fed into 2 blocks followed by a convolutional layer, a flatten layer, and an output layer. Each block comprises two convolution layers with interleaved Leaky Rectified Linear Unit (Leaky ReLU) \cite{maas2013rectifier} activation layers. These convolution layers serve a dual purpose by extracting pattern features and achieving dimensionality reduction. Meanwhile, the Leaky ReLU activation layers introduce nonlinear fitting ability and facilitate the retention of relevant features. The flatten layer is utilized to reshape the input into the vector format representing the distributions of the input patterns. 

During XGAN training, the generator produces patterns with symmetric shapes, while the discriminator computes the distribution distances between real patterns and synthetic patterns. The pre-trained F-ResNet approximates the spectral response of the generated pattern. The generator and the discriminator are trained sequentially: once the discriminator completes training on six batches, the parameter settings are updated and then fixed. Subsequently, the generator initiates training on a single batch. This training is guided by the updated discriminator as well as the pre-trained F-ResNet. Ultimately, the loss function of the generator is defined as

\begin{equation}\label{Eq:2}
\begin{split}
  L_{G} &=L_{Sinkhorn}^{G}+\lambda*L_{1}\\
&= L_{Sinkhorn}(D(x_{G}),1) + \lambda* L_{1}(c,S(x_{G})),
\end{split}
\end{equation}
where $G$, $D$, and $S$ are the generator, the discriminator, and the pre-trained surrogate model F-ResNet, respectively. $x_{G}=G(c,z)$ represents
the generated pattern $x_{G}$ by the generator with inputs of the desired response $c$ and the random Gaussian noise $z$. $L_{Sinkhorn}$ denotes the Sinkhorn distance \cite{cuturi2013sinkhorn}, and $\lambda$ is utilized to adjust the weights of losses.

The loss function of the discriminator is computed by

\begin{equation}\label{Eq:3}
\begin{split}
  L_{D} &=L_{Sinkhorn}^{D}+\gamma*L_{penalty}\\
&= L_{Sinkhorn}(D(x_{r}),-1)+L_{Sinkhorn}(D(x_{G}),1)\\
&+\gamma*L_{penalty}.
\end{split}
\end{equation}

Here, we utilize the gradient penalty $L_{penalty}$ \cite{gulrajani2017improved} to constrain the Sinkhorn distance, and $\gamma$ is utilized to adjust the weights of losses.

\section{Experiment}\label{Experiment}

\subsection{Data Collection}

\begin{figure}[htbp]
\centering
  \includegraphics[width=\linewidth]{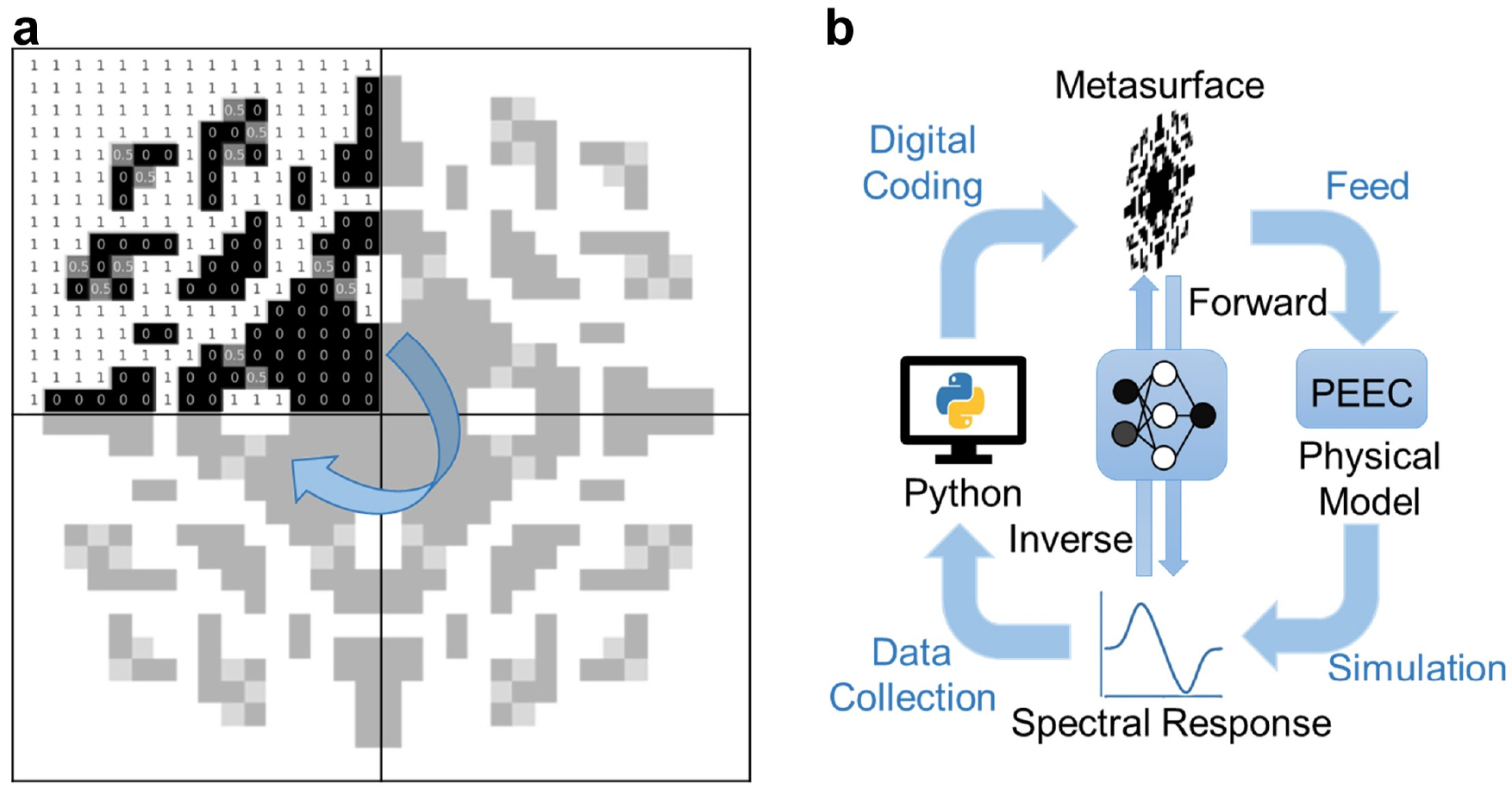}
  \caption{Data collection process. (a) Discretization and coding of the metasurface pattern. (b) The process of collecting data based on Python-PEEC co-simulation. The collected data are utilized for training neural networks that replace the manual work in forward and inverse processes.}
  \label{fig:data_collection}
\end{figure}

Fig. \ref{fig:data_collection}a illustrates the discretization and coding of metasurface patterns. The physical dimensions of a metasurface on the 2D plane are both 5.4 mm. These metasurfaces are uniformly discretized as $32 \times 32$ matrices, which are $90\degree$ rotation symmetric and exhibit free-form shapes. In order to alleviate the impact of vertex-to-vertex structure, the coding sequence of the metasurface pattern equals to a ternary matrix, where `1' means air, `0' means square metal, and an additional `0.5' represents triangular metal. We use Python-PEEC co-simulation to prepare the dataset, as illustrated in Fig. \ref{fig:data_collection}b. Python first produces a random symmetric matrix, and the partial element equivalent circuit (PEEC) method \cite{jiang2022full} then calculates the magnitude of the reflection coefficient of this matrix. We add the matrix and its corresponding response to our dataset. The magnitude of the reflection coefficient of the meta-surfaces will be sampled uniformly at 100 points between 20 and 35 GHz, whose values ranging from -1 to 1. Our dataset consists of 200,000 samples, of which 180,000 are used for training and 20,000 for testing.

\subsection{Metasurface Design}

\begin{figure}[htbp]
\centering
  \includegraphics[width=\linewidth]{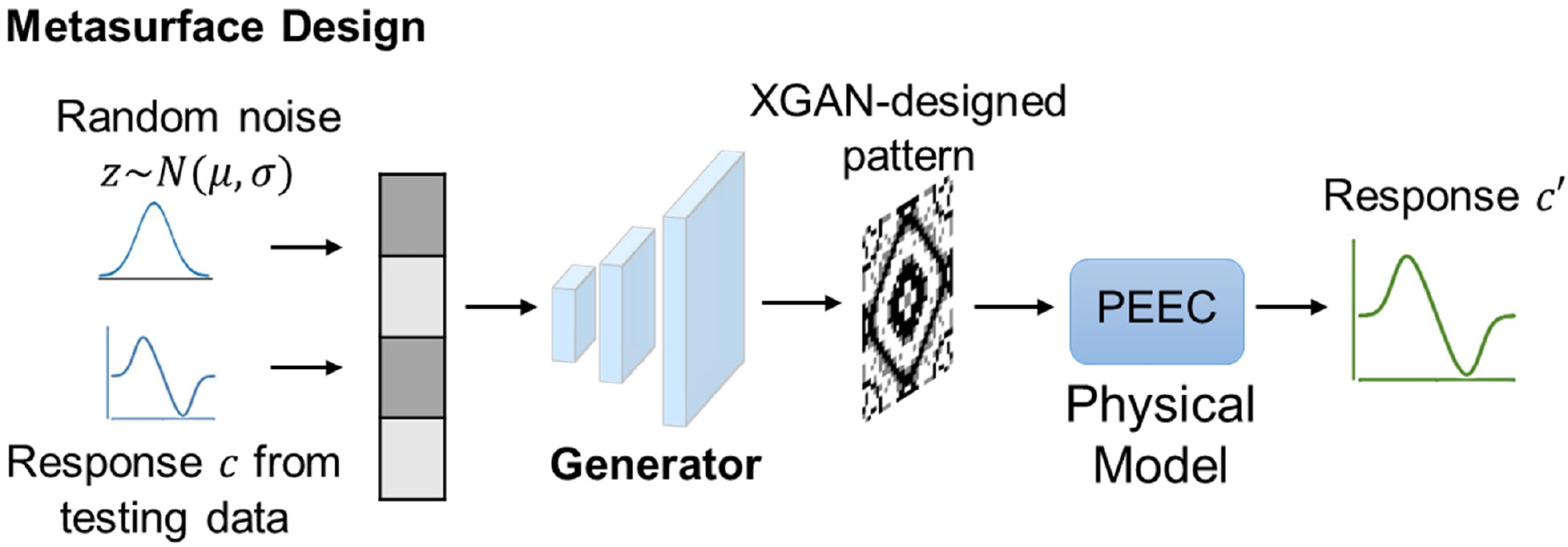}
  \caption{A sketch of metasurface design via XGAN.}
  \label{fig:metasurface_design}
\end{figure}

Fig. \ref{fig:metasurface_design} depicts the metasurface design created with XGAN. During inference, we utilize only the generator of XGAN and remove its discriminator. To generate an initial metasurface pattern, we feed a random noise following a Gaussian distribution and a target response into the generator. PEEC is adopted to calculate the response of this pattern. If the error between the calculated response and the target response is below a predetermined threshold, the initial design pattern becomes the final design.

\subsection{Implementation Details} 
The code was executed on the Ubuntu 20.04.3 LTS operating system. The deep learning algorithm was implemented using the Anaconda platform with Python 3.7.11 and PyTorch 1.7.1. These models were trained on a consumer-grade desktop that was equipped with an NVIDIA GeForce RTX 3090 GPU.

\subsection{Training Parameters} 

\begin{table*}[htbp]
	\caption{Execution details used in the training of surrogate and inverse models.}
	\centering
	\begin{tabular}[htbp]{@{}lllllll@{}}
    	\toprule
    	\multirow{2}{*}{} &CNN \cite{dai2022slmgan} & F-ResNet &WGAN-GP \cite{gulrajani2017improved}& InfoGAN \cite{chen2016infogan} &SA \cite{lee2021simulated} &  XGAN \\
	\midrule
	Software&Pytorch&Pytorch&Pytorch&Pytorch&Pytorch&Pytorch\\
	Hardware&GPU &GPU &GPU &GPU &GPU &GPU\\
    	Training set size& 180000&180000 &180000&180000&NA&180000 \\
	Test set size&20000 &20000 &20000 &20000 &20000 &20000 \\
	Optimizer & Adam&Adam &Adam&Adam&NA&Adam\\
	learning rate, $\beta$1, $\beta$2& 2e-4, 0.5, 0.999 &2e-4, 0.5, 0.999&2e-4, 0.5, 0.999&2e-4, 0.5, 0.999&NA&2e-4, 0.5, 0.999 \\
	Batch size&256&256&256&256&NA&256\\
	Epochs &1000&1000&10000&10000&1000&10000\\
	Surrogate for training &NA&NA&NA&NA&F-ResNet&F-ResNet\\
	Simulator for evaluation&NA &NA & PEEC \cite{jiang2022full} &PEEC \cite{jiang2022full} & PEEC \cite{jiang2022full}&PEEC \cite{jiang2022full}\\
\multirow{2}{*}{Evalustion Metric}& Model Fitting & Model Fitting & Model Fitting & Model Fitting & Model Fitting & Model Fitting\\
& Metrics& Metrics& Metrics& Metrics& Metrics& Metrics\\
	\bottomrule
  	\end{tabular}
	\label{Table:hyperparameters}
\end{table*}

\begin{figure*}[htbp]
\centering
  \includegraphics[width=\linewidth]{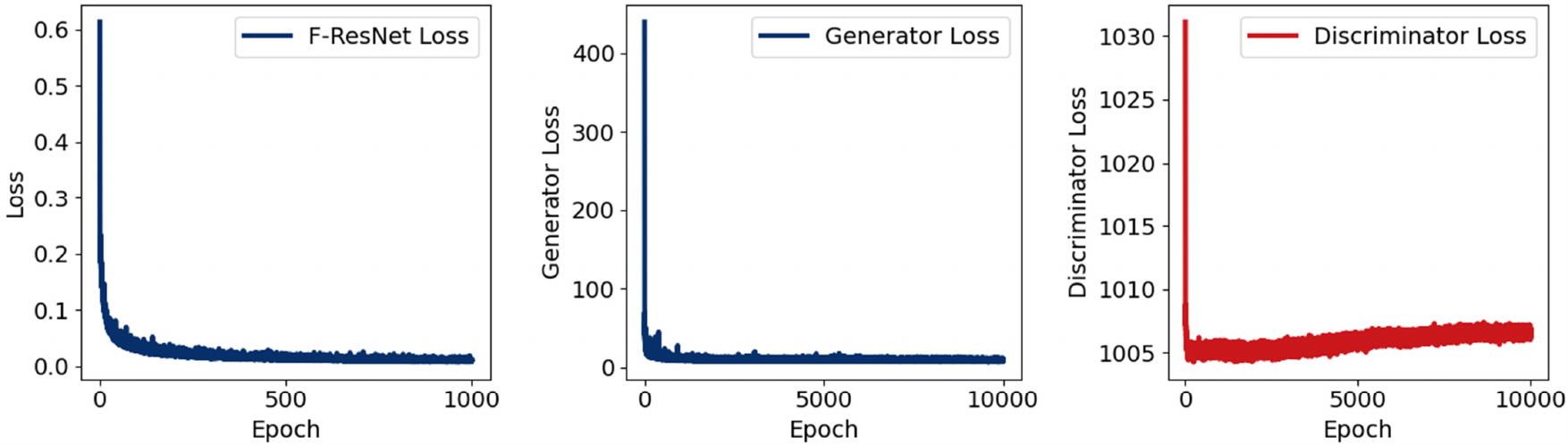}
  \caption{Training losses of the proposed models, including (a) F-ResNet loss in Eq. \ref{Eq:1}, (b) XGAN generator loss $L_{G}$ in Eq. \ref{Eq:2}, and (c) XGAN discriminator loss $L_{D}$ in Eq. \ref{Eq:3}.}
  \label{fig:XGAN_loss}
\end{figure*}

The F-ResNet and XGAN models are optimized using Adam optimizers \cite{kingma2014adam} with specific hyperparameters. The $\beta_{1}$ and $\beta_{2}$ values are set to 0.5 and 0.999, respectively, and a fixed learning rate of 0.0002 is used. Each network is equipped with its dedicated optimizer, and training is performed with a batch size of 256. The F-ResNet model is trained for a total of 1000 epochs, while the XGAN model is subjected to an extensive training regimen spanning 10000 epochs. During XGAN training, one epoch of generator training is interleaved with six epochs of discriminator training. Execution details used in the all comparative models are listed in Table \ref{Table:hyperparameters}. The training losses of F-ResNet and XGAN are depicted in Fig \ref{fig:XGAN_loss}.

\subsection{Model Fitting Metrics}

The fitting accuracy of our proposed XGAN is evaluated by four quantitative accuracy metrics $MAE_{ave}$, $ACC_{ave}$, $ACC_{min}$ and $R_{ave}^{2}$. Specially, the calculation of $ACC_{ave}$ and $ACC_{min}$ follow the equations in the reference \cite{dai2022slmgan}.  

1) Assume that the mean absolute error of each design is defined as $MAE$, the average $MAE$ of total $m$ designs is computed by:
\begin{equation}\label{Eq:6}
  MAE_{ave} = \frac{1}{m}\sum_{i=1}^{m} MAE(m).
\end{equation}

2) The average accuracy of total designs is: 
\begin{equation}\label{Eq:7}
 ACC_{ave}=1-\frac{1}{2}MAE_{ave}.
\end{equation}

3) The minimum accuracy of total designs is:

\begin{equation}\label{Eq:8}
  ACC_{min}=1-\frac{1}{2}\max_{1\leq i\leq m} {MAE(i)}.
\end{equation}

4) Average $R^{2}$ is defined as follows:

\begin{equation}\label{Eq:9}
  R^{2}_{ave} = 1-\sum_{i=1}^{m}{\frac{MSE(i)}{Var(i)}},
\end{equation}
where the $MSE$ and variance of each design are represented as $MSE$ and $Var$.

\subsection{Experimental Results}
\subsubsection{Prediction Performance for Surrogate Model F-ResNet}

\begin{figure}[htbp]
\centering
  \includegraphics[width=\linewidth]{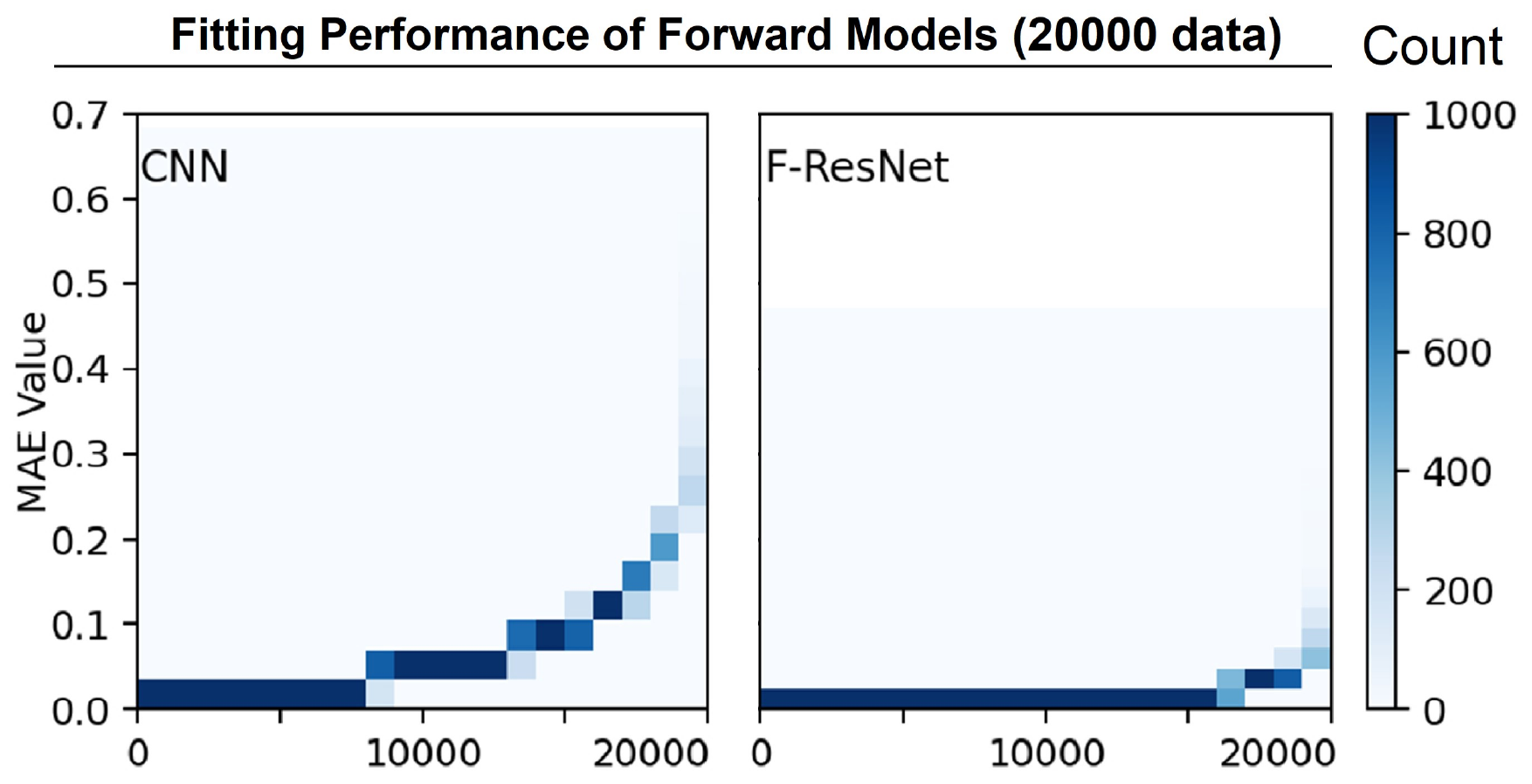}
  \caption{Density heatmaps (bins=20) of forward models (CNN, F-ResNet) comparisons regarding MAE.}
  \label{fig:forward_result}
\end{figure}

The comparisons are made to evaluate the effectiveness of the surrogate models in accurately predicting the response values of the metasurface designs. In this section, we use the CNN-based surrogate model from our previous work \cite{dai2022slmgan} as a contrastive benchmark. Density heatmaps (with 20 bins) in Fig. \ref{fig:forward_result} are used to compare the performance of surrogate models (CNN, F-ResNet) in terms of MAE values. The comparisons are based on 20,000 testing datasets. Specifically, F-ResNet achieves MAE values generally below 0.05, whereas the CNN model only achieves this level of accuracy in half of the cases (see Fig. \ref{fig:forward_result}). These results demonstrate the effectiveness of our proposed F-ResNet model for the given task. Regarding simulation times, Table \ref{Table:Accuracy_forward_inverse} reveals that both CNN and F-ResNet, as deep learning methods, have relatively quick simulation times for a given case. In contrast, PEEC, a numerical EM simulator, can only simulate 1 to 3 patterns per minute. In terms of \emph{model fitting metrics}, F-ResNet surpasses CNN in response prediction, as evidenced by superior performance in $ACC_{ave}$, $ACC_{min}$, $MAE_{ave}$, and $R^{2}$.

\begin{figure}[htbp]
\centering
  \includegraphics[width=0.8\linewidth]{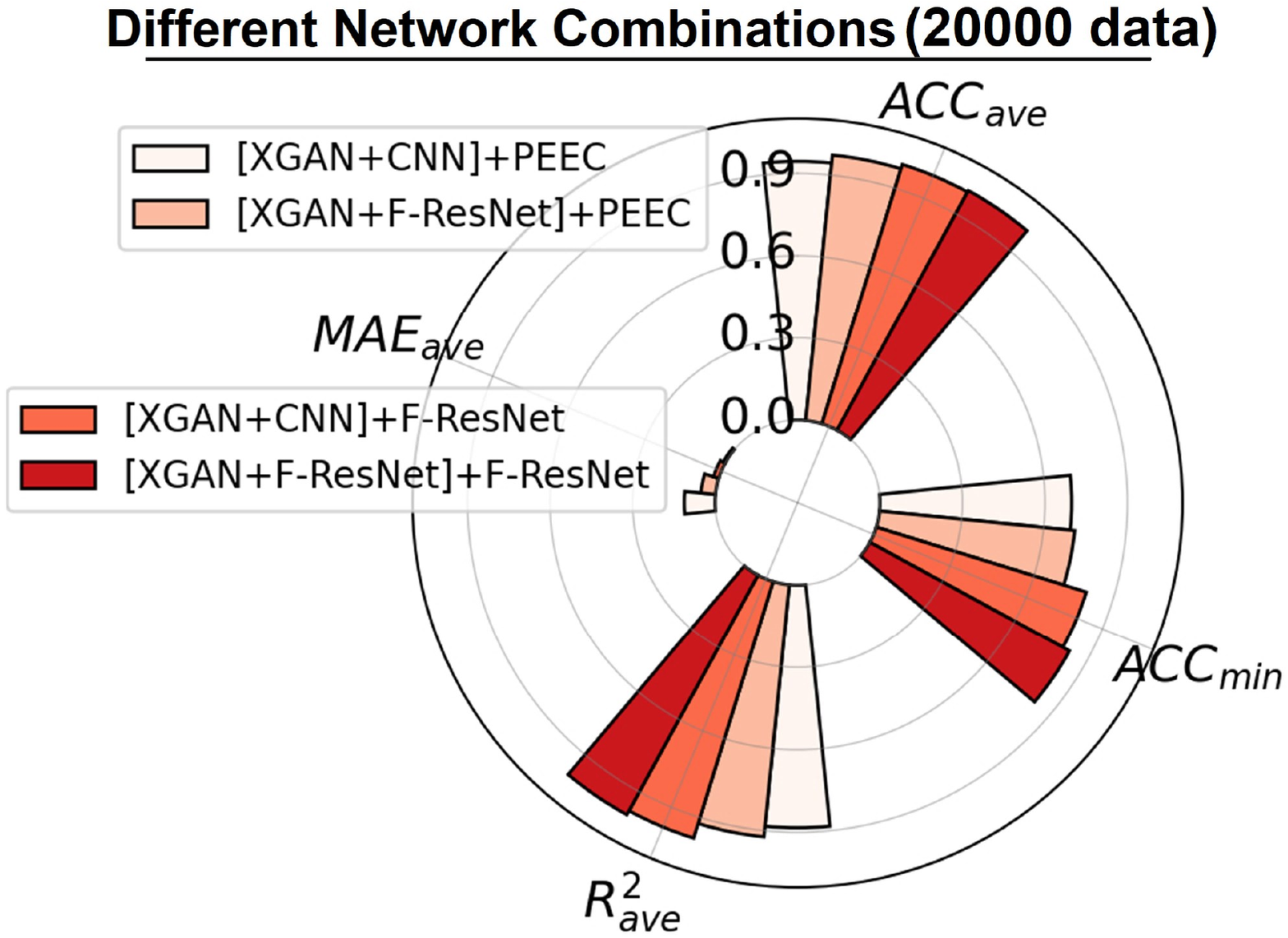}
  \caption{The polar plots of response accuracies for different network combinations (Legend meaning: [Inverse Model+Surrogate Model]+Evaluation Simulator).}
  \label{fig:combination_result}
\end{figure}

We integrate F-ResNet into various network configurations to assess its efficacy in predicting responses during XGAN training, and its performance as a simulation tool for XGAN evaluation. Our results are evaluated by using the \emph{model fitting metrics}. As presented in Fig. \ref{fig:combination_result}, incorporating F-ResNet leads to improved XGAN performance, as opposed to using CNN. This finding is validated through evaluation, where PEEC or F-ResNet is used as a simulation tool.

\begin{table*}[htbp]
	\caption{Accuracy and time cost for different forward and inverse models}
	\centering
	\begin{tabular}[htbp]{@{}lllllll@{}}
    	\toprule
     	 Model Type & Model Name & $ACC_{ave}$ & $ACC_{min}$ & $MAE_{ave}$ & $R^{2}$ & Time/Second\\
    	\cmidrule{1-7}
    	\multirow{3}{*}{Forward/Surrogate Model} & CNN \cite{dai2022slmgan} & 0.9644  & 0.6576 & 0.0712 & 0.8675 & 0.001  \\
    	& \textbf{\textcolor{ourred}{F-ResNet}}  & \textbf{\textcolor{ourred}{0.9923}}  & \textbf{\textcolor{ourred}{0.7636}} & \textbf{\textcolor{ourred}{0.0154}} & \textbf{\textcolor{ourred}{0.9869}} & \textbf{\textcolor{ourred}{0.001} }\\
    	& PEEC \cite{jiang2022full} & NA & NA & NA & NA & 20-60 \\
    	\midrule
	\multirow{4}{*}{Inverse Model} & WGAN-GP \cite{gulrajani2017improved} & 0.8221 & 0.5338 & 0.3558 & 0.7251 & 0.0015 \\
	& InfoGAN \cite{chen2016infogan} & 0.8321 & 0.6096 & 0.3358 & 0.6401 & 0.0015 \\ 
	& SA \cite{lee2021simulated} & 0.8104 & 0.5335 & 0.3791 & 0.7638 & 0.75 \\
	& \textbf{\textcolor{ourred}{XGAN}} & \textbf{\textcolor{ourred}{0.9734}} & \textbf{\textcolor{ourred}{0.7132}} & \textbf{\textcolor{ourred}{0.0533}} & \textbf{\textcolor{ourred}{0.9228}} & \textbf{\textcolor{ourred}{0.0015}} \\
	\bottomrule
  	\end{tabular}
	\label{Table:Accuracy_forward_inverse}
\end{table*}

\subsubsection{XGAN Outperforms Existing Inverse Design Methods}

The experimental approaches for evaluating inverse design in this study include WGAN-GP \cite{gulrajani2017improved}, InfoGAN \cite{chen2016infogan}, the simulated annealing (SA) \cite{lee2021simulated}, and XGAN.

\begin{figure*}[htbp]
\centering
  \includegraphics[width=0.9\linewidth]{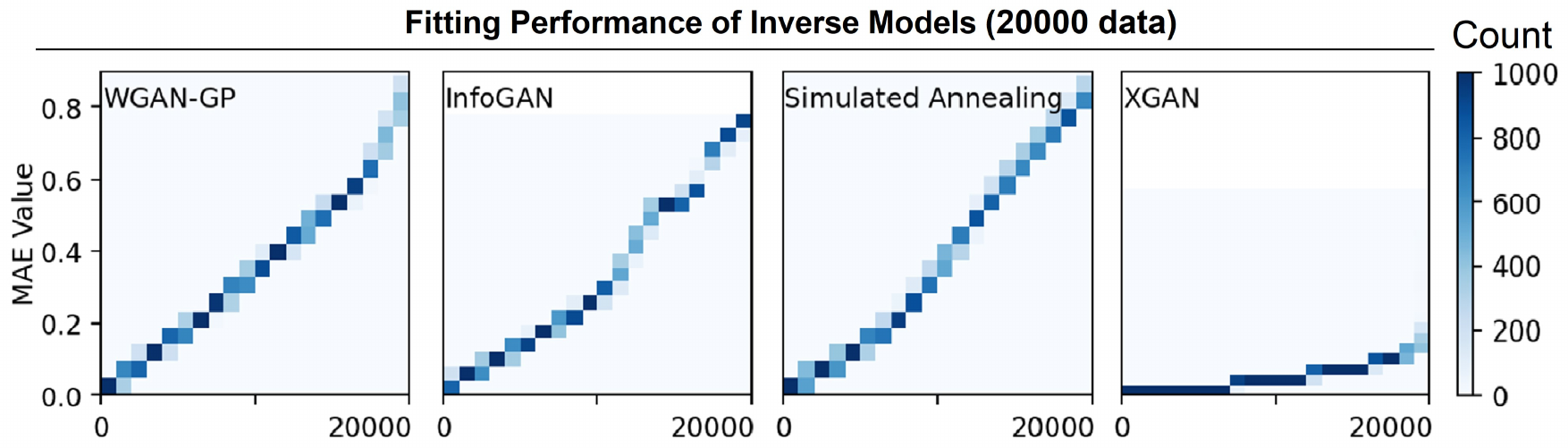}
  \caption{Density heatmaps (bins=20) of response MAE for inverse models (WGAN-GP, InfoGAN, Simulated Annealing, and XGAN).}
  \label{fig:XGAN_result}
\end{figure*}

\textbf{Quantitative Comparison:} Fig. \ref{fig:XGAN_result} highlights the superior performance of XGAN compared to other contrastive methods regarding MAE. Specifically, the MAE values achieved by XGAN are generally below 0.1, whereas WGAN-GP and the SA have the highest MAE value of 1, and InfoGAN has the highest value of 0.8. Table \ref{Table:Accuracy_forward_inverse} provides the running time of each inverse model. During inference, the proposed XGAN model takes 0.0015 seconds to generate a solution for a given response.

\begin{figure*}[htbp]
\centering
  \includegraphics[width=0.9\linewidth]{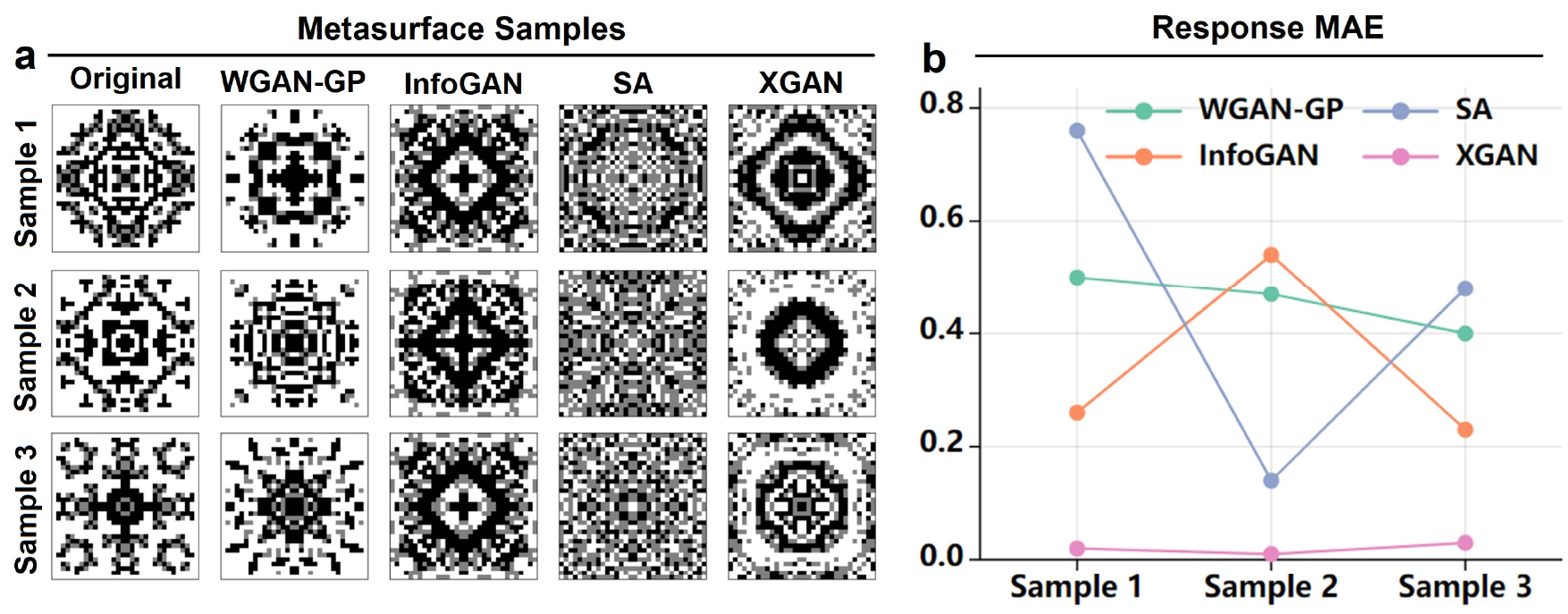}
  \caption{Free-form metasurface design by various inverse models, including WGAN-GP, InfoGAN, Simulated Annealing (SA), and XGAN. (a) Model sample and original data visualization, when Gaussian noise is $N(\mu=0, \sigma=1)$ and input responses are variable. (b) The MAE between the input response and the PEEC-simulated response of each sample.}
  \label{fig:XGAN_sample}
\end{figure*}

\textbf{Qualitative Comparison:} Fig. \ref{fig:XGAN_sample}a provides some metasurface designs encoded by WGAN-GP, InfoGAN, SA, and XGAN. These models are fed with the target responses, and the corresponding designed patterns are simulated by PEEC to obtain the actual responses. Our aim is to determine which model could generate the highest-quality patterns that accurately matched the target responses. Specifically, XGAN generates patterns with response MAEs below 0.1, whereas the other models fail to achieve the same level of accuracy (see Fig. \ref{fig:XGAN_sample}b). This finding indicates the proficiency of XGAN in generating patterns with desired responses. This superior performance can be attributed to the use of interpretable latent vectors from the XGAN and the response feedback mechanism from F-ResNet, which allow for condition control and precise refinement of the generated patterns. To sum up, XGAN showcases the distinctive ability to generate appropriate patterns based on input responses, setting it apart from other models that lack this unique capability.

\section{Conclusion}

In this paper, we present XGAN, a novel approach for the rapid and accurate realization of free-form metasurfaces. We combine ResNet50 with a multi-head attention model, named F-ResNet, to accurately predict the responses of metasurface patterns. F-ResNet exhibits impressive performance, achieving a prediction rate of 1000 patterns per second, with $ACC_{ave}$ of 0.99 and $R^{2}$ of 0.98. This is 20000-60000 times faster than the conventional simulator PEEC, effectively replacing it as a surrogate simulation tool. XGAN includes a GAN for inverse design. To address the limitation of traditional GAN discriminators in capturing response relevance between original and generated patterns, we integrate F-ResNet into the framework. F-ResNet acts as a constraint on XGAN training by connecting it with the generator. During the training, XGAN generates an initial pattern using input vectors, and iteratively refines it based on feedback from F-ResNet, ultimately achieving the desired response. To assess the efficiency of XGAN, we design \emph{model fitting metrics}. XGAN achieves $ACC_{ave}$ of 0.97 for 20000 designs in a single run, and produces a metasurface pattern in 0.0015 seconds, which is 500 times faster than the conventional optimization method SA. Overall, XGAN provides a fast and accurate approach for designing free-form metasurfaces to meet the response requirement. This makes XGAN suitable for various applications, including light manipulation and spectral selection.

\bibliographystyle{IEEEtran}
\bibliography{sn-bibliography}

\begin{thebibliography}{10}
\providecommand{\url}[1]{#1}
\csname url@samestyle\endcsname
\providecommand{\newblock}{\relax}
\providecommand{\bibinfo}[2]{#2}
\providecommand{\BIBentrySTDinterwordspacing}{\spaceskip=0pt\relax}
\providecommand{\BIBentryALTinterwordstretchfactor}{4}
\providecommand{\BIBentryALTinterwordspacing}{\spaceskip=\fontdimen2\font plus
\BIBentryALTinterwordstretchfactor\fontdimen3\font minus
  \fontdimen4\font\relax}
\providecommand{\BIBforeignlanguage}[2]{{%
\expandafter\ifx\csname l@#1\endcsname\relax
\typeout{** WARNING: IEEEtran.bst: No hyphenation pattern has been}%
\typeout{** loaded for the language `#1'. Using the pattern for}%
\typeout{** the default language instead.}%
\else
\language=\csname l@#1\endcsname
\fi
#2}}
\providecommand{\BIBdecl}{\relax}
\BIBdecl

\bibitem{yang2021demonstration}
Y.~Yang, Y.~Ge, R.~Li, X.~Lin, D.~Jia, Y.-j. Guan, S.-q. Yuan, H.-x. Sun,
  Y.~Chong, and B.~Zhang, ``Demonstration of negative refraction induced by
  synthetic gauge fields,'' \emph{Science Advances}, vol.~7, no.~50, p.
  eabj2062, 2021.

\bibitem{cheng2023large}
H.-Y. Cheng, M.-J. Ye, W.-R. Chen, C.-Y. Yang, S.-W. Chu, K.-P. Chen, and K.-H.
  Lin, ``Large optical modulation of dielectric huygens’ metasurface
  absorber,'' \emph{Advanced Optical Materials}, p. 2300102, 2023.

\bibitem{shaltout2019spatiotemporal}
A.~M. Shaltout, V.~M. Shalaev, and M.~L. Brongersma, ``Spatiotemporal light
  control with active metasurfaces,'' \emph{Science}, vol. 364, no. 6441, p.
  eaat3100, 2019.

\bibitem{chen2020microwave}
Z.~N. Chen, T.~Li, and W.~E. Liu, ``Microwave metasurface-based lens antennas
  for 5g and beyond,'' in \emph{2020 14th European Conference on Antennas and
  Propagation (EuCAP)}.\hskip 1em plus 0.5em minus 0.4em\relax IEEE, 2020, pp.
  1--4.

\bibitem{cui2014coding}
T.~J. Cui, M.~Q. Qi, X.~Wan, J.~Zhao, and Q.~Cheng, ``Coding metamaterials,
  digital metamaterials and programmable metamaterials,'' \emph{Light: science
  \& applications}, vol.~3, no.~10, pp. e218--e218, 2014.

\bibitem{wan2016field}
X.~Wan, M.~Q. Qi, T.~Y. Chen, and T.~J. Cui, ``Field-programmable beam
  reconfiguring based on digitally-controlled coding metasurface,''
  \emph{Scientific reports}, vol.~6, no.~1, p. 20663, 2016.

\bibitem{qiu2019deep}
T.~Qiu, X.~Shi, J.~Wang, Y.~Li, S.~Qu, Q.~Cheng, T.~Cui, and S.~Sui, ``Deep
  learning: a rapid and efficient route to automatic metasurface design,''
  \emph{Advanced Science}, vol.~6, no.~12, p. 1900128, 2019.

\bibitem{zhu2021phase}
R.~Zhu, T.~Qiu, J.~Wang, S.~Sui, C.~Hao, T.~Liu, Y.~Li, M.~Feng, A.~Zhang,
  C.-W. Qiu \emph{et~al.}, ``Phase-to-pattern inverse design paradigm for fast
  realization of functional metasurfaces via transfer learning,'' \emph{Nature
  communications}, vol.~12, no.~1, p. 2974, 2021.

\bibitem{li2017electromagnetic}
L.~Li, T.~Jun~Cui, W.~Ji, S.~Liu, J.~Ding, X.~Wan, Y.~Bo~Li, M.~Jiang, C.-W.
  Qiu, and S.~Zhang, ``Electromagnetic reprogrammable coding-metasurface
  holograms,'' \emph{Nature communications}, vol.~8, no.~1, p. 197, 2017.

\bibitem{jia2023knowledge}
Y.~Jia, C.~Qian, Z.~Fan, T.~Cai, E.-P. Li, and H.~Chen, ``A knowledge-inherited
  learning for intelligent metasurface design and assembly,'' \emph{Light:
  Science \& Applications}, vol.~12, no.~1, p.~82, 2023.

\bibitem{liu2022programmable}
C.~Liu, Q.~Ma, Z.~J. Luo, Q.~R. Hong, Q.~Xiao, H.~C. Zhang, L.~Miao, W.~M. Yu,
  Q.~Cheng, L.~Li \emph{et~al.}, ``A programmable diffractive deep neural
  network based on a digital-coding metasurface array,'' \emph{Nature
  Electronics}, vol.~5, no.~2, pp. 113--122, 2022.

\bibitem{zhu2023metasurfaces}
Y.~Zhu, X.~Zang, H.~Chi, Y.~Zhou, Y.~Zhu, and S.~Zhuang, ``Metasurfaces
  designed by a bidirectional deep neural network and iterative algorithm for
  generating quantitative field distributions,'' \emph{Light: Advanced
  Manufacturing}, vol.~4, no.~1, pp. 1--11, 2023.

\bibitem{zhang2022heterogeneous}
J.~Zhang, C.~Qian, Z.~Fan, J.~Chen, E.~Li, J.~Jin, and H.~Chen, ``Heterogeneous
  transfer-learning-enabled diverse metasurface design,'' \emph{Advanced
  Optical Materials}, vol.~10, no.~17, p. 2200748, 2022.

\bibitem{jia2022situ}
Y.~Jia, C.~Qian, Z.~Fan, Y.~Ding, Z.~Wang, D.~Wang, E.-P. Li, B.~Zheng, T.~Cai,
  and H.~Chen, ``In situ customized illusion enabled by global metasurface
  reconstruction,'' \emph{Advanced Functional Materials}, vol.~32, no.~19, p.
  2109331, 2022.

\bibitem{nadell2019deep}
C.~C. Nadell, B.~Huang, J.~M. Malof, and W.~J. Padilla, ``Deep learning for
  accelerated all-dielectric metasurface design,'' \emph{Optics express},
  vol.~27, no.~20, pp. 27\,523--27\,535, 2019.

\bibitem{li2022empowering}
Z.~Li, R.~Pestourie, Z.~Lin, S.~G. Johnson, and F.~Capasso, ``Empowering
  metasurfaces with inverse design: principles and applications,'' \emph{ACS
  Photonics}, vol.~9, no.~7, pp. 2178--2192, 2022.

\bibitem{tanriover2022deep}
I.~Tanriover, D.~Lee, W.~Chen, and K.~Aydin, ``Deep generative modeling and
  inverse design of manufacturable free-form dielectric metasurfaces,''
  \emph{ACS Photonics}, 2022.

\bibitem{wang2022deep}
J.~Wang, R.~Xi, T.~Cai, H.~Lu, R.~Zhu, B.~Zheng, and H.~Chen, ``Deep neural
  network with data cropping algorithm for absorptive frequency-selective
  transmission metasurface,'' \emph{Advanced Optical Materials}, vol.~10,
  no.~13, p. 2200178, 2022.

\bibitem{xiong2019controlling}
B.~Xiong, L.~Deng, R.~Peng, and Y.~Liu, ``Controlling the degrees of freedom in
  metasurface designs for multi-functional optical devices,'' \emph{Nanoscale
  Advances}, vol.~1, no.~10, pp. 3786--3806, 2019.

\bibitem{liu2021tackling}
Z.~Liu, D.~Zhu, L.~Raju, and W.~Cai, ``Tackling photonic inverse design with
  machine learning,'' \emph{Advanced Science}, vol.~8, no.~5, p. 2002923, 2021.

\bibitem{zhang2021deep}
T.~Zhang, C.~Y. Kee, Y.~S. Ang, and L.~Ang, ``Deep learning-based design of
  broadband ghz complex and random metasurfaces,'' \emph{APL Photonics},
  vol.~6, no.~10, p. 106101, 2021.

\bibitem{yun2022deep}
J.~Yun, S.~Kim, S.~So, M.~Kim, and J.~Rho, ``Deep learning for topological
  photonics,'' \emph{Advances in Physics: X}, vol.~7, no.~1, p. 2046156, 2022.

\bibitem{mall2020cyclical}
A.~Mall, A.~Patil, A.~Sethi, and A.~Kumar, ``A cyclical deep learning based
  framework for simultaneous inverse and forward design of nanophotonic
  metasurfaces,'' \emph{Scientific reports}, vol.~10, no.~1, pp. 1--12, 2020.

\bibitem{hodge2019multi}
J.~A. Hodge, K.~V. Mishra, and A.~I. Zaghloul, ``Multi-discriminator
  distributed generative model for multi-layer rf metasurface discovery,'' in
  \emph{2019 IEEE Global Conference on Signal and Information Processing
  (GlobalSIP)}.\hskip 1em plus 0.5em minus 0.4em\relax IEEE, 2019, pp. 1--5.

\bibitem{creswell2018generative}
A.~Creswell, T.~White, V.~Dumoulin, K.~Arulkumaran, B.~Sengupta, and A.~A.
  Bharath, ``Generative adversarial networks: An overview,'' \emph{IEEE signal
  processing magazine}, vol.~35, no.~1, pp. 53--65, 2018.

\bibitem{wang2021generative}
Z.~Wang, Q.~She, and T.~E. Ward, ``Generative adversarial networks in computer
  vision: A survey and taxonomy,'' \emph{ACM Computing Surveys (CSUR)},
  vol.~54, no.~2, pp. 1--38, 2021.

\bibitem{kingma2014stochastic}
D.~P. Kingma and M.~Welling, ``Stochastic gradient vb and the variational
  auto-encoder,'' in \emph{Second international conference on learning
  representations, ICLR}, vol.~19, 2014, p. 121.

\bibitem{liu2018generative}
Z.~Liu, D.~Zhu, S.~P. Rodrigues, K.-T. Lee, and W.~Cai, ``Generative model for
  the inverse design of metasurfaces,'' \emph{Nano letters}, vol.~18, no.~10,
  pp. 6570--6576, 2018.

\bibitem{ma2019probabilistic}
W.~Ma, F.~Cheng, Y.~Xu, Q.~Wen, and Y.~Liu, ``Probabilistic representation and
  inverse design of metamaterials based on a deep generative model with
  semi-supervised learning strategy,'' \emph{Advanced Materials}, vol.~31,
  no.~35, p. 1901111, 2019.

\bibitem{naseri2020machine}
P.~Naseri and S.~V. Hum, ``A machine learning-based approach to synthesize
  multilayer metasurfaces,'' in \emph{2020 IEEE International Symposium on
  Antennas and Propagation and North American Radio Science Meeting}.\hskip 1em
  plus 0.5em minus 0.4em\relax IEEE, 2020, pp. 933--934.

\bibitem{clerc2010particle}
M.~Clerc, \emph{Particle swarm optimization}.\hskip 1em plus 0.5em minus
  0.4em\relax John Wiley \& Sons, 2010, vol.~93.

\bibitem{an2021multifunctional}
S.~An, B.~Zheng, H.~Tang, M.~Y. Shalaginov, L.~Zhou, H.~Li, M.~Kang, K.~A.
  Richardson, T.~Gu, J.~Hu \emph{et~al.}, ``Multifunctional metasurface design
  with a generative adversarial network,'' \emph{Advanced Optical Materials},
  vol.~9, no.~5, p. 2001433, 2021.

\bibitem{dai2022slmgan}
M.~Dai, Y.~Jiang, F.~Yang, X.~Xu, W.~Zhao, M.~H. Dao, and Y.~Liu, ``Slmgan:
  Single-layer metasurface design with symmetrical free-form patterns using
  generative adversarial networks,'' \emph{Applied Soft Computing}, vol. 130,
  p. 109646, 2022.

\bibitem{an2020deep}
S.~An, B.~Zheng, M.~Y. Shalaginov, H.~Tang, H.~Li, L.~Zhou, J.~Ding, A.~M.
  Agarwal, C.~Rivero-Baleine, M.~Kang \emph{et~al.}, ``Deep learning modeling
  approach for metasurfaces with high degrees of freedom,'' \emph{Optics
  Express}, vol.~28, no.~21, pp. 31\,932--31\,942, 2020.

\bibitem{yang2022ultraspectral}
J.~Yang, K.~Cui, X.~Cai, J.~Xiong, H.~Zhu, S.~Rao, S.~Xu, Y.~Huang, F.~Liu,
  X.~Feng \emph{et~al.}, ``Ultraspectral imaging based on metasurfaces with
  freeform shaped meta-atoms,'' \emph{Laser \& Photonics Reviews}, vol.~16,
  no.~7, p. 2100663, 2022.

\bibitem{yla2003calculation}
P.~Yla-Oijala and M.~Taskinen, ``Calculation of cfie impedance matrix elements
  with rwg and n/spl times/rwg functions,'' \emph{IEEE Transactions on Antennas
  and Propagation}, vol.~51, no.~8, pp. 1837--1846, 2003.

\bibitem{He2015}
K.~He, X.~Zhang, S.~Ren, and J.~Sun, ``Deep residual learning for image
  recognition,'' \emph{arXiv preprint arXiv:1512.03385}, 2015.

\bibitem{vaswani2017attention}
A.~Vaswani, N.~Shazeer, N.~Parmar, J.~Uszkoreit, L.~Jones, A.~N. Gomez,
  {\L}.~Kaiser, and I.~Polosukhin, ``Attention is all you need,''
  \emph{Advances in neural information processing systems}, vol.~30, 2017.

\bibitem{radford2021learning}
A.~Radford, J.~W. Kim, C.~Hallacy, A.~Ramesh, G.~Goh, S.~Agarwal, G.~Sastry,
  A.~Askell, P.~Mishkin, J.~Clark \emph{et~al.}, ``Learning transferable visual
  models from natural language supervision,'' in \emph{International conference
  on machine learning}.\hskip 1em plus 0.5em minus 0.4em\relax PMLR, 2021, pp.
  8748--8763.

\bibitem{jing2022neural}
G.~Jing, P.~Wang, H.~Wu, J.~Ren, Z.~Xie, J.~Liu, H.~Ye, Y.~Li, D.~Fan, and
  S.~Chen, ``Neural network-based surrogate model for inverse design of
  metasurfaces,'' \emph{Photonics Research}, vol.~10, no.~6, pp. 1462--1471,
  2022.

\bibitem{tanriover2020physics}
I.~Tanriover, W.~Hadibrata, and K.~Aydin, ``Physics-based approach for a neural
  networks enabled design of all-dielectric metasurfaces,'' \emph{ACS
  Photonics}, vol.~7, no.~8, pp. 1957--1964, 2020.

\bibitem{naseri2021generative}
P.~Naseri and S.~V. Hum, ``A generative machine learning-based approach for
  inverse design of multilayer metasurfaces,'' \emph{IEEE Transactions on
  Antennas and Propagation}, vol.~69, no.~9, pp. 5725--5739, 2021.

\bibitem{zelaszczyk2023cross}
M.~{\.Z}elaszczyk and J.~Ma{\'n}dziuk, ``Cross-modal text and visual
  generation: A systematic review. part 1—image to text,'' \emph{Information
  Fusion}, 2023.

\bibitem{ramos2023smallcap}
R.~Ramos, B.~Martins, D.~Elliott, and Y.~Kementchedjhieva, ``Smallcap:
  lightweight image captioning prompted with retrieval augmentation,'' in
  \emph{Proceedings of the IEEE/CVF Conference on Computer Vision and Pattern
  Recognition}, 2023, pp. 2840--2849.

\bibitem{couairon2022embedding}
G.~Couairon, M.~Douze, M.~Cord, and H.~Schwenk, ``Embedding arithmetic of
  multimodal queries for image retrieval,'' in \emph{Proceedings of the
  IEEE/CVF Conference on Computer Vision and Pattern Recognition}, 2022, pp.
  4950--4958.

\bibitem{yao2023detclipv2}
L.~Yao, J.~Han, X.~Liang, D.~Xu, W.~Zhang, Z.~Li, and H.~Xu, ``Detclipv2:
  Scalable open-vocabulary object detection pre-training via word-region
  alignment,'' in \emph{Proceedings of the IEEE/CVF Conference on Computer
  Vision and Pattern Recognition}, 2023, pp. 23\,497--23\,506.

\bibitem{gao2022transform}
F.~Gao, Q.~Ping, G.~Thattai, A.~Reganti, Y.~N. Wu, and P.~Natarajan,
  ``Transform-retrieve-generate: Natural language-centric outside-knowledge
  visual question answering,'' in \emph{Proceedings of the IEEE/CVF Conference
  on Computer Vision and Pattern Recognition}, 2022, pp. 5067--5077.

\bibitem{kil2023prestu}
J.~Kil, S.~Changpinyo, X.~Chen, H.~Hu, S.~Goodman, W.-L. Chao, and R.~Soricut,
  ``Prestu: Pre-training for scene-text understanding,'' in \emph{Proceedings
  of the IEEE/CVF International Conference on Computer Vision}, 2023, pp.
  15\,270--15\,280.

\bibitem{gulrajani2017improved}
I.~Gulrajani, F.~Ahmed, M.~Arjovsky, V.~Dumoulin, and A.~C. Courville,
  ``Improved training of wasserstein gans,'' \emph{Advances in neural
  information processing systems}, vol.~30, 2017.

\bibitem{he2015delving}
K.~He, X.~Zhang, S.~Ren, and J.~Sun, ``Delving deep into rectifiers: Surpassing
  human-level performance on imagenet classification,'' in \emph{Proceedings of
  the IEEE international conference on computer vision}, 2015, pp. 1026--1034.

\bibitem{shi2016real}
W.~Shi, J.~Caballero, F.~Husz{\'a}r, J.~Totz, A.~P. Aitken, R.~Bishop,
  D.~Rueckert, and Z.~Wang, ``Real-time single image and video super-resolution
  using an efficient sub-pixel convolutional neural network,'' in
  \emph{Proceedings of the IEEE conference on computer vision and pattern
  recognition}, 2016, pp. 1874--1883.

\bibitem{maas2013rectifier}
A.~L. Maas, A.~Y. Hannun, A.~Y. Ng \emph{et~al.}, ``Rectifier nonlinearities
  improve neural network acoustic models,'' in \emph{Proc. icml}, vol.~30,
  no.~1.\hskip 1em plus 0.5em minus 0.4em\relax Atlanta, GA, 2013, p.~3.

\bibitem{cuturi2013sinkhorn}
M.~Cuturi, ``Sinkhorn distances: Lightspeed computation of optimal transport,''
  \emph{Advances in neural information processing systems}, vol.~26, 2013.

\bibitem{jiang2022full}
Y.~Jiang, W.-J. Zhao, R.~X.-K. Gao, E.-X. Liu, and C.~E. Png, ``A full-wave
  generalized peec model for periodic metallic structure with arbitrary
  shape,'' \emph{IEEE Transactions on Microwave Theory and Techniques},
  vol.~70, no.~9, pp. 4110--4119, 2022.

\bibitem{chen2016infogan}
X.~Chen, Y.~Duan, R.~Houthooft, J.~Schulman, I.~Sutskever, and P.~Abbeel,
  ``Infogan: Interpretable representation learning by information maximizing
  generative adversarial nets,'' \emph{Advances in neural information
  processing systems}, vol.~29, 2016.

\bibitem{lee2021simulated}
J.~Lee and D.~Perkins, ``A simulated annealing algorithm with a dual
  perturbation method for clustering,'' \emph{Pattern Recognition}, vol. 112,
  p. 107713, 2021.

\bibitem{kingma2014adam}
D.~P. Kingma and J.~Ba, ``Adam: A method for stochastic optimization,''
  \emph{arXiv preprint arXiv:1412.6980}, 2014.

\end{thebibliography}



\begin{IEEEbiography}[{\includegraphics[width=1in,height=1.25in,clip,keepaspectratio]{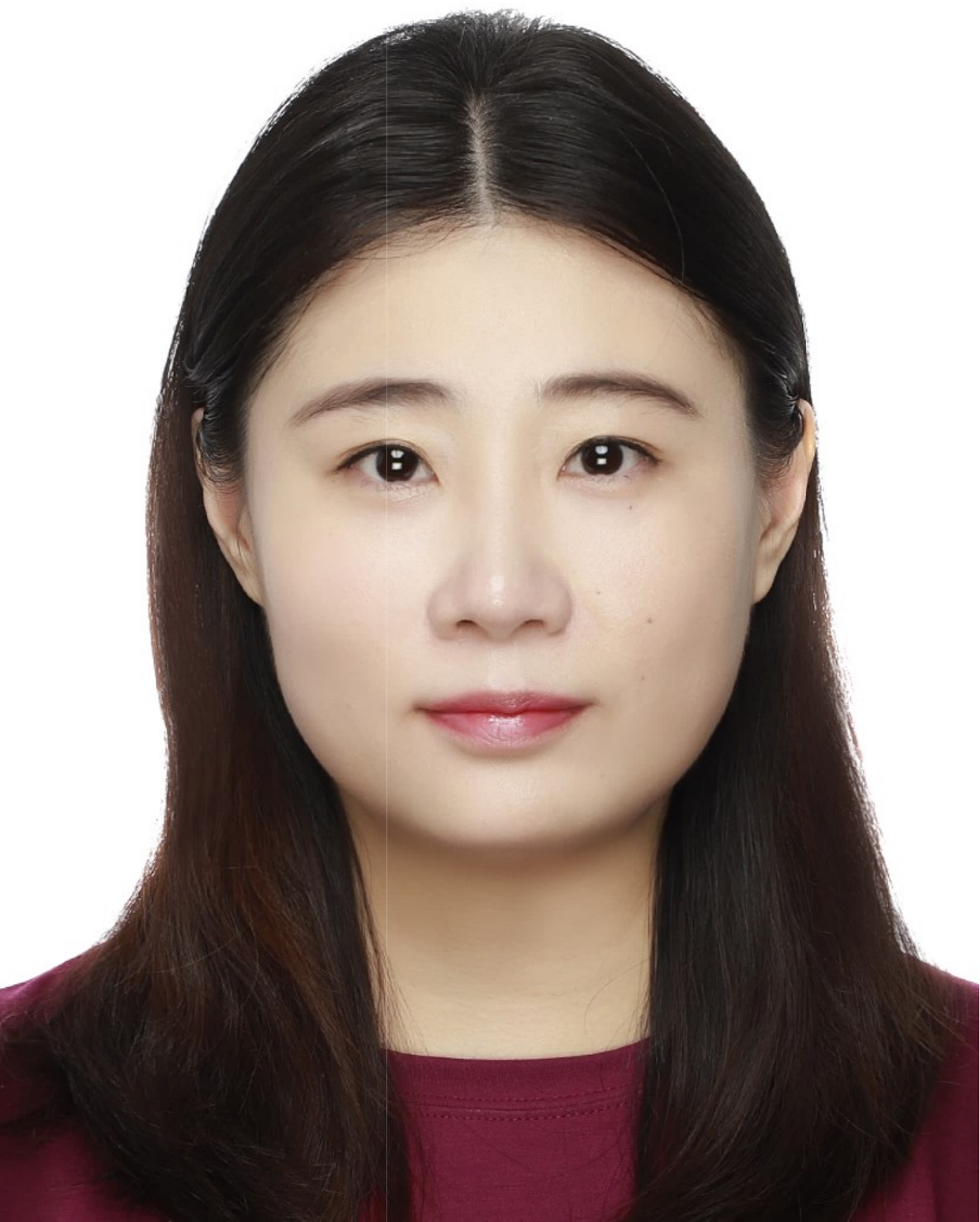}}]{Manna Dai}
received her Ph.D. degree in Computer Science from Faculty of Engineering and Information Technology, University of Technology Sydney, Australia. She was affiliated with Data61, Commonwealth Scientific and Industrial ResearchOrganization, Australia. She conducted her postdoctoral research in the Division of Engineering in Medicine, Department of Medicine, Brigham and Women’s Hospital, Harvard Medical School, USA. Currently, she works as a Scientist in Computing \&Intelligence Department, Institute of High Performance Computing, Agency for Science, Technology and Research (A*STAR), Singapore. Her research interests include inverse design, target tracking, machine learning, deep learning, computer vision, and image-processing.
\end{IEEEbiography}

\vspace{1pt}

\begin{IEEEbiography}[{\includegraphics[width=1in,height=1.25in,clip,keepaspectratio]{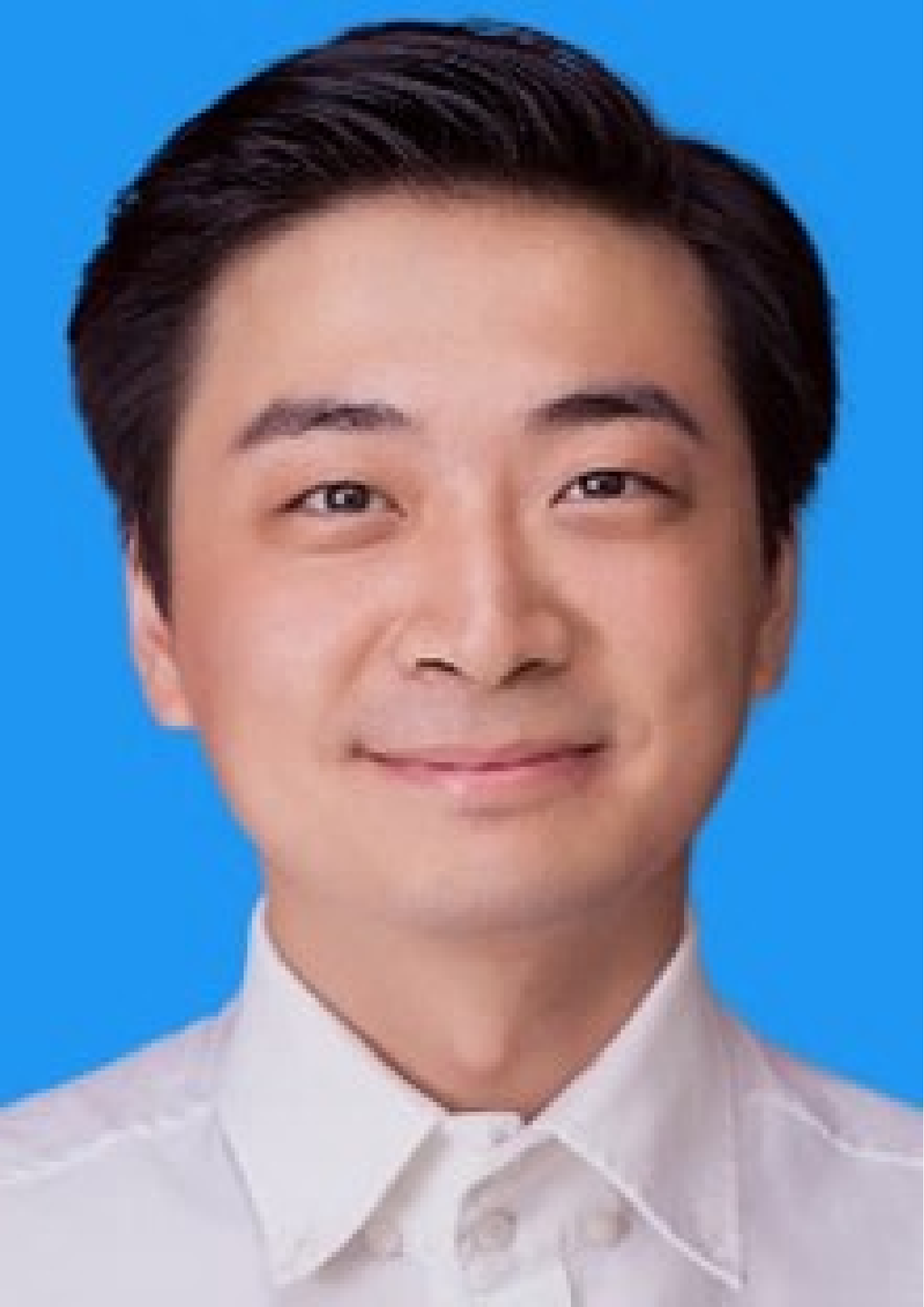}}]{Yang Jiang}
(Member, IEEE) was born in Shandong, China in 1990. He received the B.S. and the Ph.D. degrees in electronic engineering from the Chinese University of Hong Kong, Hong Kong, China, in 2013 and 2019.
He is currently a Research Scientist with the Institute of High-Performance Computing (IHPC), A*STAR, Singapore. His research interests include partial element equivalent circuit (PEEC) modeling for cutting-edge technologies, physics-informed artificial intelligence for inverse design and inverse scattering problems, and EMC/ESD modeling. He received the Highly Recommend Paper Award of APEMC in 2022. He serves as an Executive Committee Member of IEEE Singapore EMC Chapter.
\end{IEEEbiography}

\vspace{1pt}

\begin{IEEEbiography}[{\includegraphics[width=1in,height=1.25in,clip,keepaspectratio]{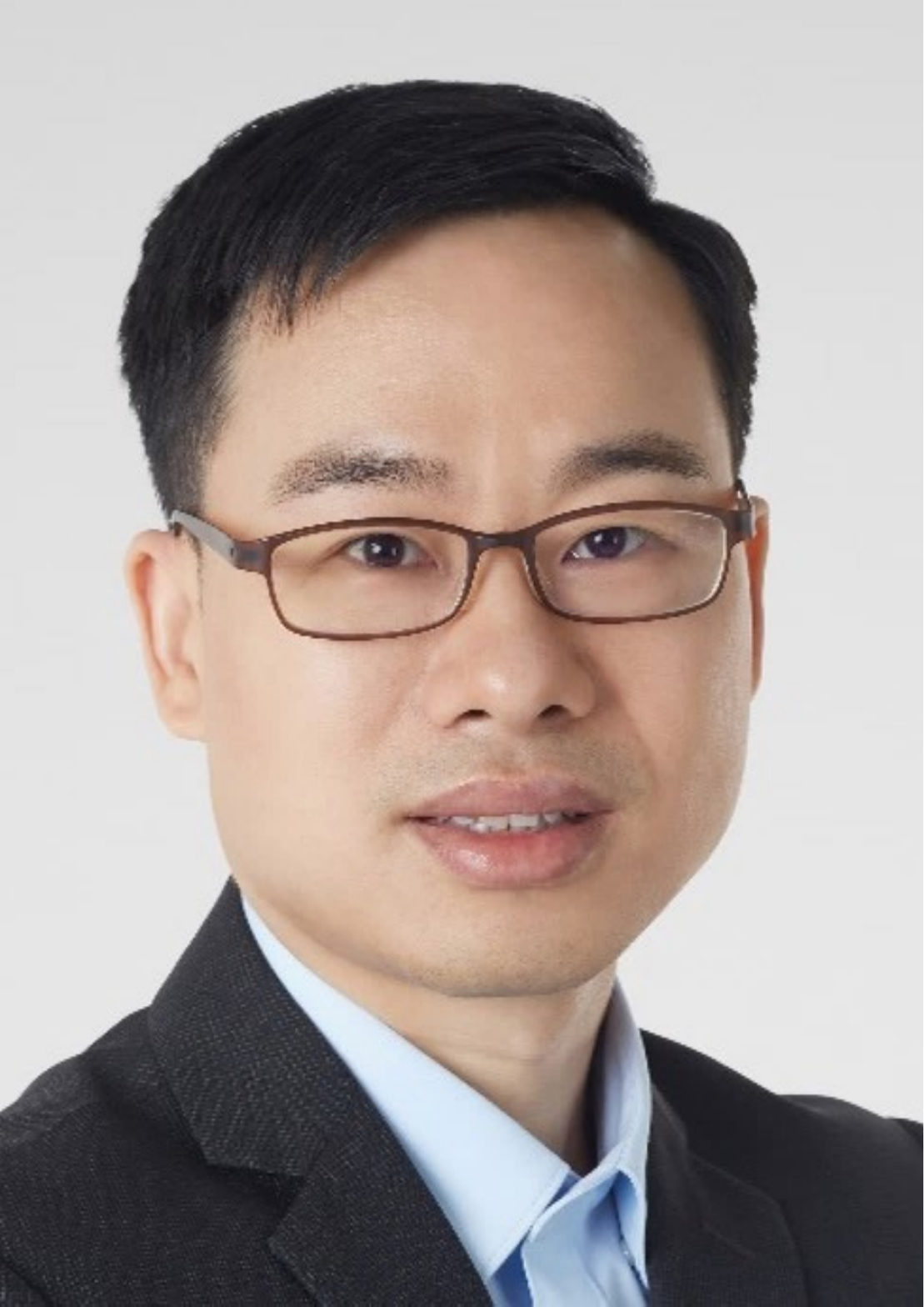}}]{Feng Yang}
received his B.Eng. degree in information engineering and M.Eng. degree in control engineering from Xi’an Jiaotong University (XJTU), Xi’an, China. He then received his Ph.D. degree in machine learning and bioinformatics from Nanyang Technological University (NTU), Singapore, in 2012.
From June 2011 to March 2012, he was a Research Associate at NTU. In 2012, he joined the Institute of High Performance Computing (IHPC), Agency for Science, Technology, and Research (A*STAR), Singapore. Currently, he is a Principal Scientist in the Department of Computing \& Intelligence, IHPC. His research interests include machine learning, knowledge-driven AI, generative AI for inverse problems, information retrieval, representation learning, data-based diagnosis and prognosis, and digital twin.
\end{IEEEbiography}

\vspace{1pt}

\begin{IEEEbiography}[{\includegraphics[width=1in,height=1.25in,clip,keepaspectratio]{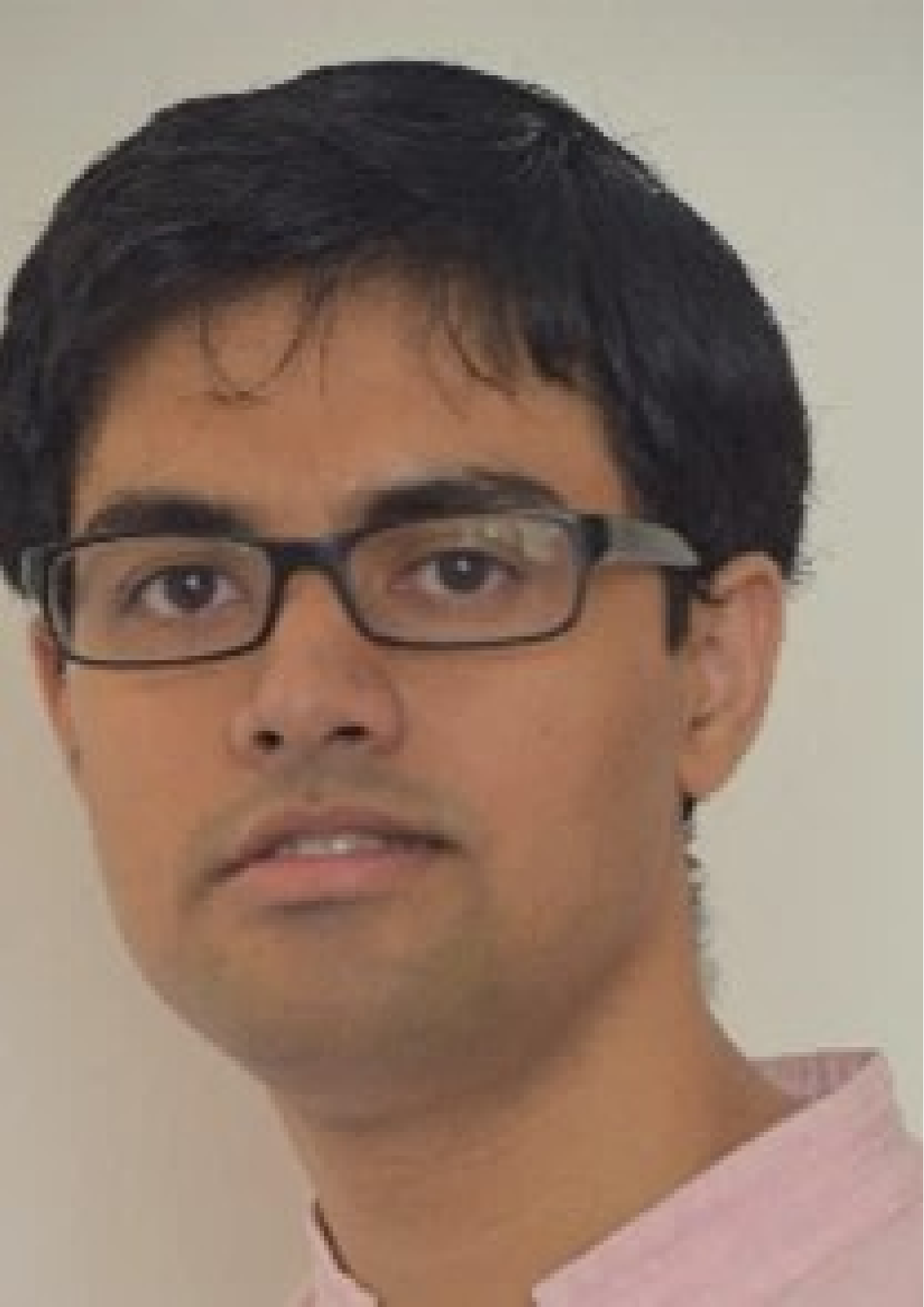}}]{Joyjit Chattoraj}
is a senior scientist at Institute of High Performance Computing (IHPC), Agency for Science Technology and Research (A*STAR), Singapore. He received his PhD degree in Physics from Université Paris Est, France. His research interests include Physics-based Artificial Intelligence, Inverse Modelling, and Artificial Intelligence for Material Discovery.  
\end{IEEEbiography}

\vspace{1pt}

\begin{IEEEbiography}[{\includegraphics[width=1in,height=1.25in,clip,keepaspectratio]{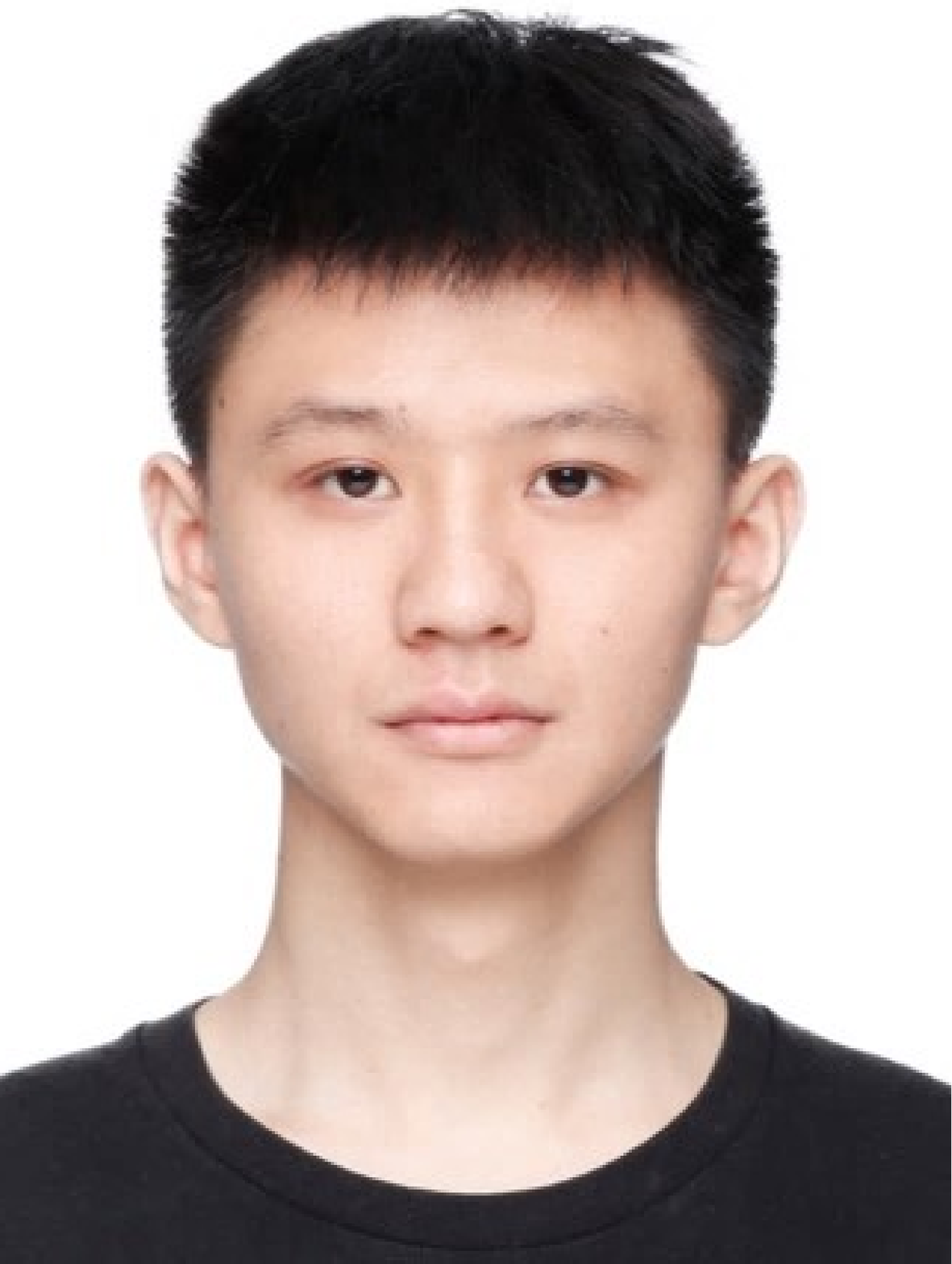}}]{Yingzhi Xia}
received the B.S. degree in Mathematics and Applied Mathematics from Northwestern Polytechnical University, China, and the Ph.D. degree in Computer Science from the University of Chinese Academy of Sciences, China, and ShanghaiTech University, China. He is currently a research scientist at the Institute of High Performance Computing (IHPC), A*STAR, Singapore. His research interests include inverse problems, probabilistic machine learning, and scientific computing. 
\end{IEEEbiography}

\vspace{1pt}

\begin{IEEEbiography}[{\includegraphics[width=1in,height=1.25in,clip,keepaspectratio]{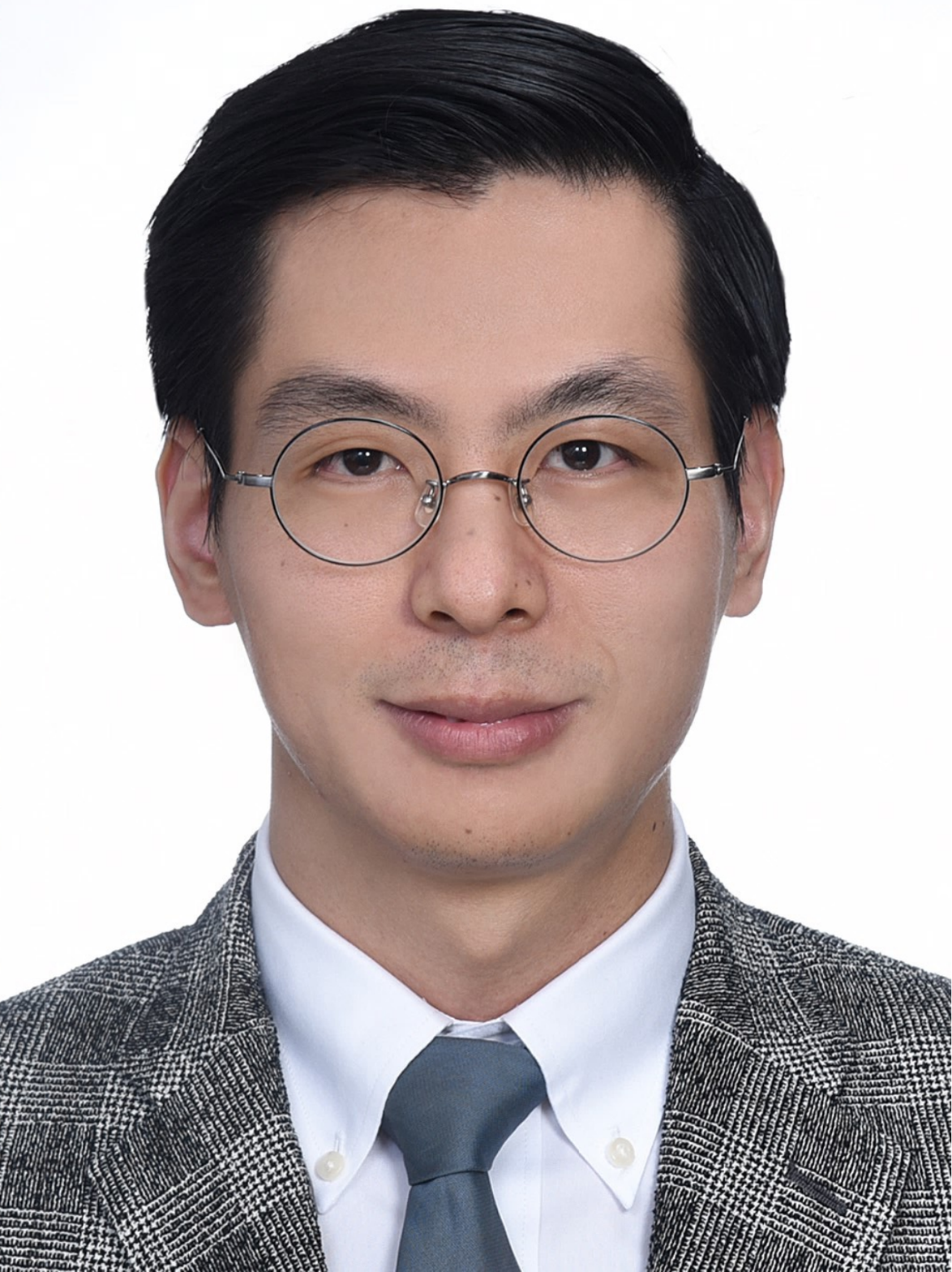}}]{Xinxing Xu}
(Member, IEEE) is a senior scientist and Group Manager of Multimodal AI, Computing \& Intelligence Department at IHPC, A*STAR. He obtained his Ph.D. in Computer Engineering from Nanyang Technological University (NTU), Singapore and his bachelor's degree from the University of Science and Technology of China (USTC). 

His research interests include machine learning, computer vision, and medical data analysis. He has published research works in top-tier AI journals and conferences including IEEE TPAMI, IEEE TNNLS, IEEE TIP, IEEE JBHI, CVPR, ICCV, IJCAI, ICDM, ECCV, and MICCAI. A few of his recent research works on deep learning for medical imaging have also been published in top-tier venues including the New England Journal of Medicine, Nature Medicine, Nature Aging, Nature Machine Intelligence, and The Lancet Digital Health. He received Best Paper Winner at OMIA workshop of MICCAI 2022 and Best Paper Award at BeyondLabeler Workshop at IJCAI 2016. He also won 3rd Place in Glaucoma grAding from Multi-Modality imAges (GAMMA) Challenge at MICCAI 2021.
\end{IEEEbiography}

\vspace{1pt}

\begin{IEEEbiography}[{\includegraphics[width=1in,height=1.25in,clip,keepaspectratio]{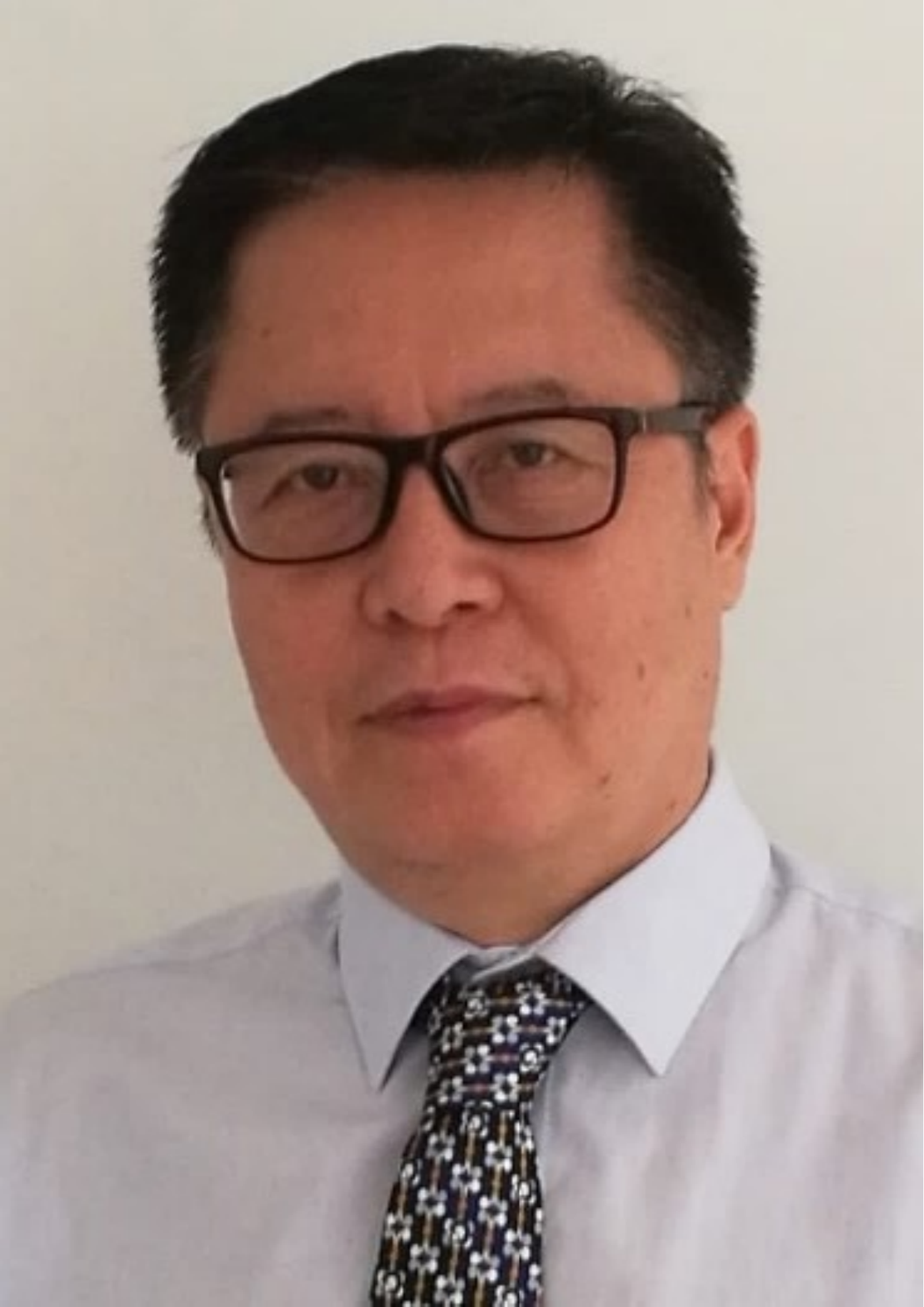}}]{Weijiang Zhao}
received his bachelor degree from Nankai University in 1989, the M.Eng and the Ph.D. from Xidian University in 1992 and 1999, respectively. 
He was with Xidian University as an Associate Professor from 1999 to 2001. He then joined the National University of Singapore as a Research Scientist.  He has been with A*STAR Institute of High Performance Computing, Singapore since 2010, and is currently a Senior Principal Scientist in the Electronics and Photonics Department.  He was part of the project team responsible for the Integrated Environmental Modeller software which won the President’s Technology Award in 2019.
Dr Zhao has authored and coauthored more than 60 technical papers. His research interests include computational electromagnetic, inverse design of metasurfaces with AI, antennas and propagation, electromagnetic compatibility and interference, radome analysis, electromagnetic scattering, wireless communications, traffic noise modeling, GNSS positioning, ionospheric scintillation, numerical methods, and optimization techniques.
\end{IEEEbiography}

\vspace{1pt}

\begin{IEEEbiography}[{\includegraphics[width=1in,height=1.25in,clip,keepaspectratio]{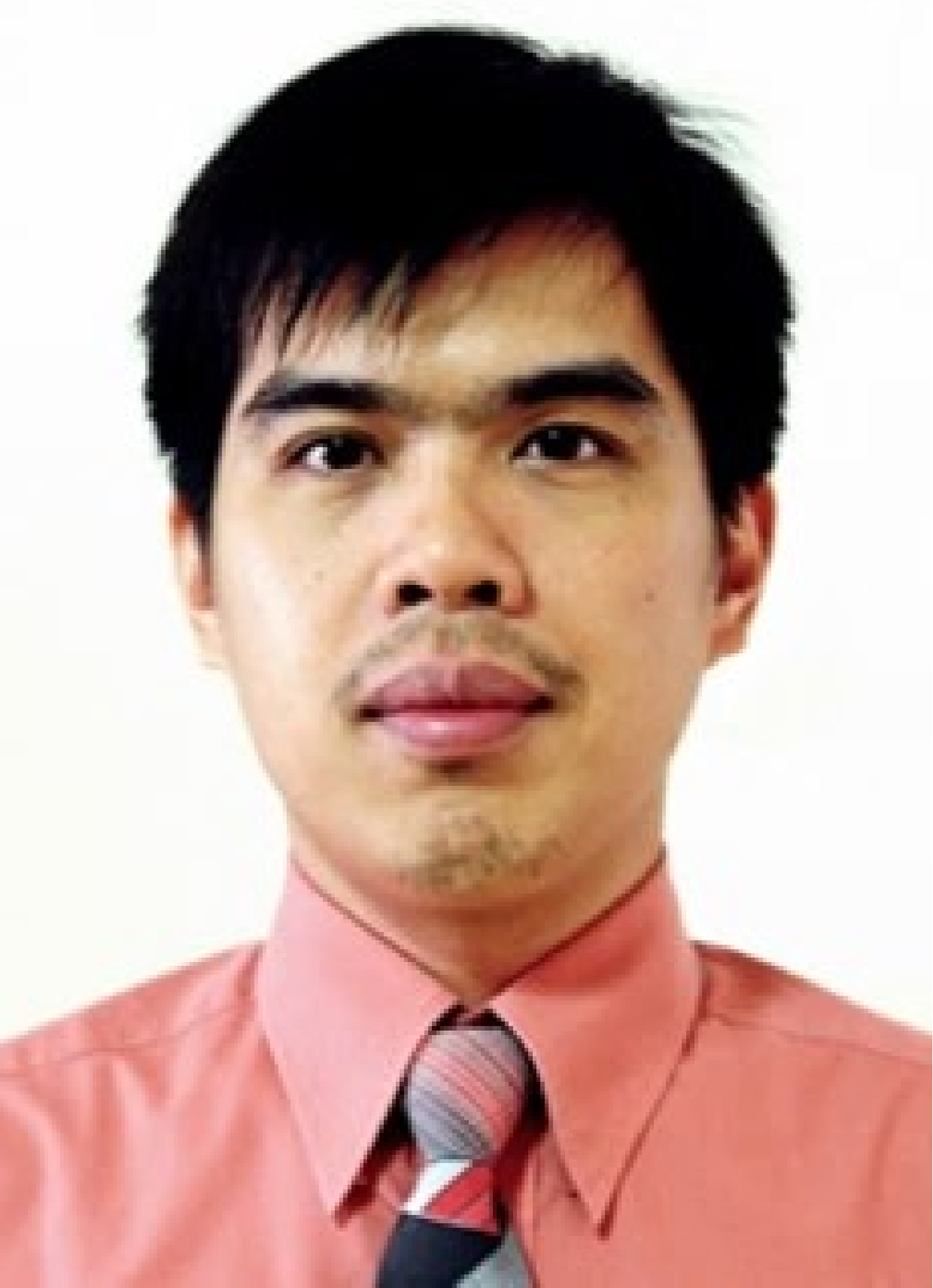}}]{My Ha Dao}
received the B.S. degree in Aeronautical Engineering from the Ho Chi Minh City University of Technology in 2002, and the M.Eng. degree in High Performance Computing for Engineered Systems and the PhD degree in Civil Engineering from the National University of Singapore in 2004 and 2011, respectively.
He work for 7 years at the Tropical Marine Science Institute and 3 years at the Singapore-MIT Alliance for Research and Technology, National University of Singapore. He is currently a Principal Scientist at the Institute Of High Performance Computing (IHPC), A*STAR. His research interests include physics-based, data driven and artificial intelligence models that complement high fidelity computational numerical models in aerospace, marine and offshore, additive manufacturing, optics applications.
\end{IEEEbiography}

\vspace{1pt}

\begin{IEEEbiography}[{\includegraphics[width=1in,height=1.25in,clip,keepaspectratio]{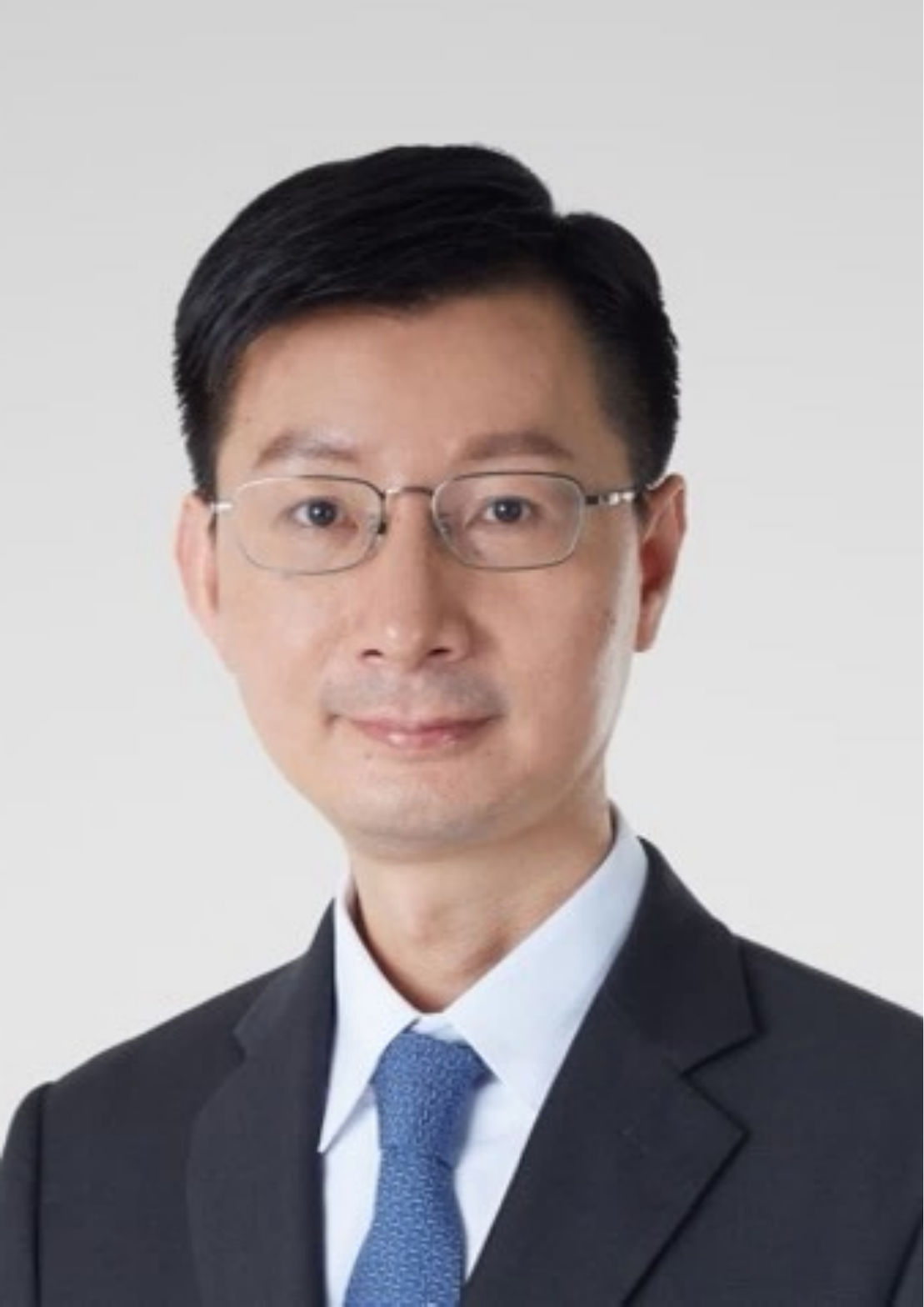}}]{Yong Liu}
Deputy Department Director, Computing \& Intelligence Department at Institute of High Performance Computing (IHPC), A*STAR, Singapore. He is also Adjunct Associate Professor at Duke-NUS Medical School, NUS and Adjunct Principal Investigator at Singapore Eye Research Institute (SERI). 
Being passionate on the potential of AI on healthcare, he has led multiple research projects in multimodal machine learning, medical imaging analysis, especially AI for digital ophthalmology and other areas in healthcare. Together with clinicians, he has been awarded large scale programme and multiple grants as PI in the area of AI for healthcare. He and his team have been awarded multiple grants from AI SG, NMRC, IAF-PP, and NHIC. He and his team have also published top tier papers in New England Journal of Medicine, Lancet Digital Health, Nature Aging, Ophthalmology, Neurology, Annals of Neurology, and also top-tier AI papers (AAAI, CVPR, IJCAI, MICCAI). He has won the Nation Award (Covid-19) Commendation Medal. His papers received Best Paper Award at REMIA workshop with 2022 MICCAI, Best Paper Award at BeyondLabeler Workshop on International Joint Conference on Artificial Intelligence (IJCAI) 2016. He and his team has also won the 3rd Place GAMMA Challenge in MICCAI 2021, the NSCC Outstanding HPC Innovation Award 2017, and the first prize of Rakuten TV recommendation Challenge 2015. 
\end{IEEEbiography}

\vfill

\end{document}